\documentclass[10pt,journal,compsoc]{IEEEtran}

\ifCLASSOPTIONcompsoc
  \usepackage[nocompress]{cite}
\else
  \usepackage{cite}
\fi

\usepackage{amsmath,amssymb,graphicx}
\usepackage{amsfonts}
\usepackage{url}
\usepackage{algorithm}
\usepackage{algorithmicx}
\usepackage{algpseudocode}
\usepackage{booktabs}
\usepackage{threeparttable}
\usepackage{multirow}
\usepackage{subcaption}
\usepackage{caption}
\usepackage{array}
\usepackage{color}

\graphicspath{{figure/}}

\begin{document}

\title{GPCA: A Probabilistic Framework for Gaussian Process Embedded Channel Attention}

\author{Jiyang~Xie,~\IEEEmembership{Student Member,~IEEE,}
        Zhanyu~Ma,~\IEEEmembership{Senior Member,~IEEE,}\\
        Dongliang~Chang,
        Guoqiang~Zhang,~\IEEEmembership{Member,~IEEE,}
        and~Jun~Guo
\IEEEcompsocitemizethanks{\IEEEcompsocthanksitem J. Xie, Z. Ma, D. Chang, and J. Guo are with the Pattern Recognition and Intelligent System Lab., School of Artificial Intelligence, Beijing University of Posts and Telecommunications, China. E-mail: $\{$xiejiyang$2013$, mazhanyu, changdongliang, guojun$\}$@bupt.edu.cn\protect\\
\IEEEcompsocthanksitem G. Zhang is with the School of Electrical and Data Engineering, University of Technology Sydney, Australia. E-mail: guoqiang.zhang@uts.edu.au\protect\\}
\thanks{
(Corresponding author: Zhanyu Ma)}}

\markboth{Journal of \LaTeX\ Class Files,~Vol.~xx, No.~x, ~20xx}%
{Shell \MakeLowercase{\textit{et al.}}: Bare Demo of IEEEtran.cls for Computer Society Journals}

\IEEEtitleabstractindextext{%
\begin{abstract}
    Channel attention mechanisms have been commonly applied in many visual tasks for effective performance improvement. It is able to reinforce the informative channels as well as to suppress the useless channels. Recently, different channel attention modules have been proposed and implemented in various ways. Generally speaking, they are mainly based on convolution and pooling operations. In this paper, we propose Gaussian process embedded channel attention (GPCA) module and further interpret the channel attention schemes in a probabilistic way. The GPCA module intends to model the correlations among the channels, which are assumed to be captured by beta distributed variables. As the beta distribution cannot be integrated into the end-to-end training of convolutional neural networks (CNNs) with a mathematically tractable solution, we utilize an approximation of the beta distribution to solve this problem. To specify, we adapt a Sigmoid-Gaussian approximation, in which the Gaussian distributed variables are transferred into the interval $[0,1]$. The Gaussian process is then utilized to model the correlations among different channels. In this case, a mathematically tractable solution is derived. The GPCA module can be efficiently implemented and integrated into the end-to-end training of the CNNs. Experimental results demonstrate the promising performance of the proposed GPCA module. Codes are available at~\url{https://github.com/PRIS-CV/GPCA}.
\end{abstract}

}

\maketitle

\IEEEdisplaynontitleabstractindextext

\IEEEpeerreviewmaketitle

\section{Introduction}\label{sec:introduction}

\IEEEPARstart{D}{eep} neural networks have attracted great attention and improved the performance of various computer vision tasks from academic to industry including image classification~\cite{simonyan2015very,he2016deep,huang2017densely,peng2019few}, object detection~\cite{liu2016ssd}, semantic segmentation~\cite{chen2018deeplab,yuan2019object}, and image retrieval~\cite{li2019deep}. Convolutional neural networks (CNNs) as well-known deep neural network architectures are able to effectively capture the discriminative nonlinear patterns from images, extract feature representations, and generate multi-channel feature maps. These phenomena are mainly due to their deep and wide structures, and the introduction of the additional components including dropout~\cite{hinton12} and attention schemes~\cite{vaswani2017attention}.

\begin{figure}[!t]
  \includegraphics[width=0.9\linewidth]{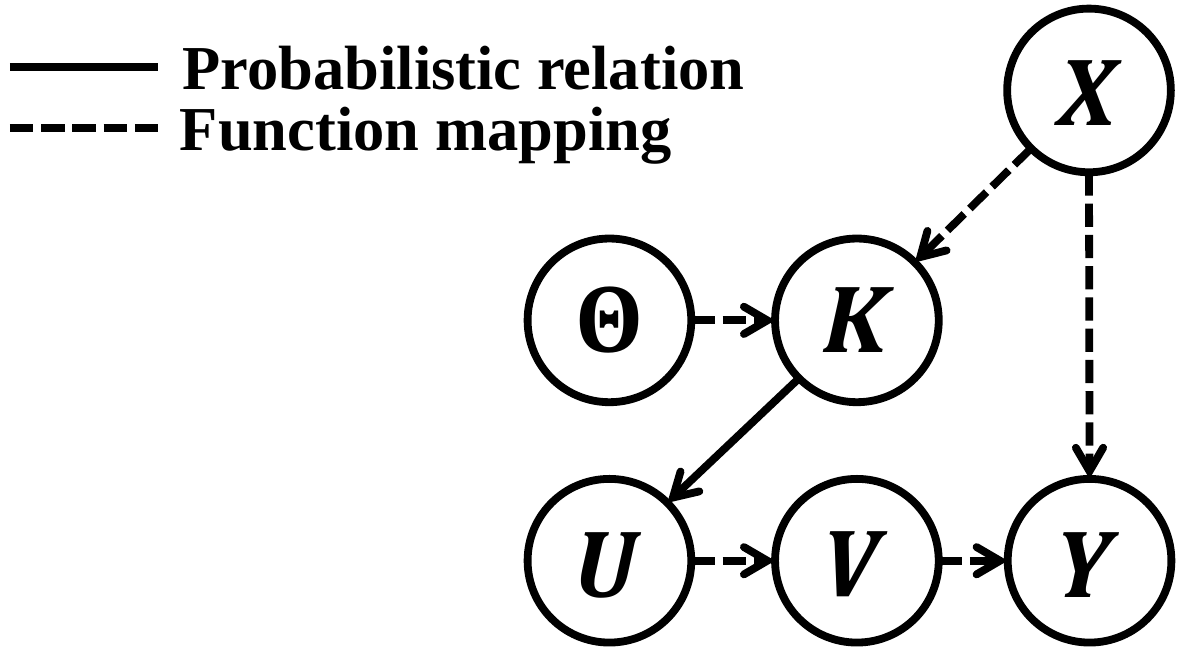}
  \caption{Probabilistic graphical model of the Gaussian process embedded channel attention (GPCA) module. $\boldsymbol{X}\in R^{C\times W\times H}$ are the input feature maps, where $C$, $W$, and $H$ are the number of channels, the width, and the height, respectively. Meanwhile, the final output $\boldsymbol{Y}\in R^{C\times W\times H}$ of the GPCA module is obtained by scaling $\boldsymbol{X}$ with the attention mask vector $\boldsymbol{V}\in R^{C}$. Each element in $\boldsymbol{V}$ is generated by the corresponding elements in $\boldsymbol{U}\in R^{C}$. Then, it is transferred and bounded into the interval $[0,1]$. $\boldsymbol{U}$ is the channel weight vector, optimized by a Gaussian process (GP) with the Gram matrix $\boldsymbol{K}\in R^{C\times C}$. $\boldsymbol{K}$ directly represents the channel correlations in $\boldsymbol{X}$ and is calculated by a kernel function with parameters $\boldsymbol{\Theta}$.}\label{fig:graphicalmodel}
  \vspace{-6mm}
\end{figure}

Attention mechanisms as the key components of CNN architectures are generally located behind specific convolutional layers, such as the last layers in each residual block of ResNet~\cite{he2016deep}. They have been utilized for feature recalibration with attention weights and widely applied to effectively improve the performance of the CNNs learned from large-scale datasets~\cite{hu2018squeeze}. Previous work~\cite{jaderberg2015spatial,woo2018cbam,hu2018squeeze} have intensively studied the significance of the attention mechanisms. The attention scheme can not only inform the CNNs about their concentrations on an image or the extracted features, but also develop the feature representation of the image~\cite{woo2018cbam}.

Channel attention, which is a significant part of the attention mechanisms, aims to improve the quality of feature representations by precisely modeling the correlations between the channels~\cite{hu2018squeeze}. It is able to enhance the informative and discriminative feature maps and suppress the useless and unhelpful ones simultaneously by learning the attention weights for each channel, which are optimized implicitly.

\begin{figure*}[!t]
  \centering
  \includegraphics[width=0.95\linewidth]{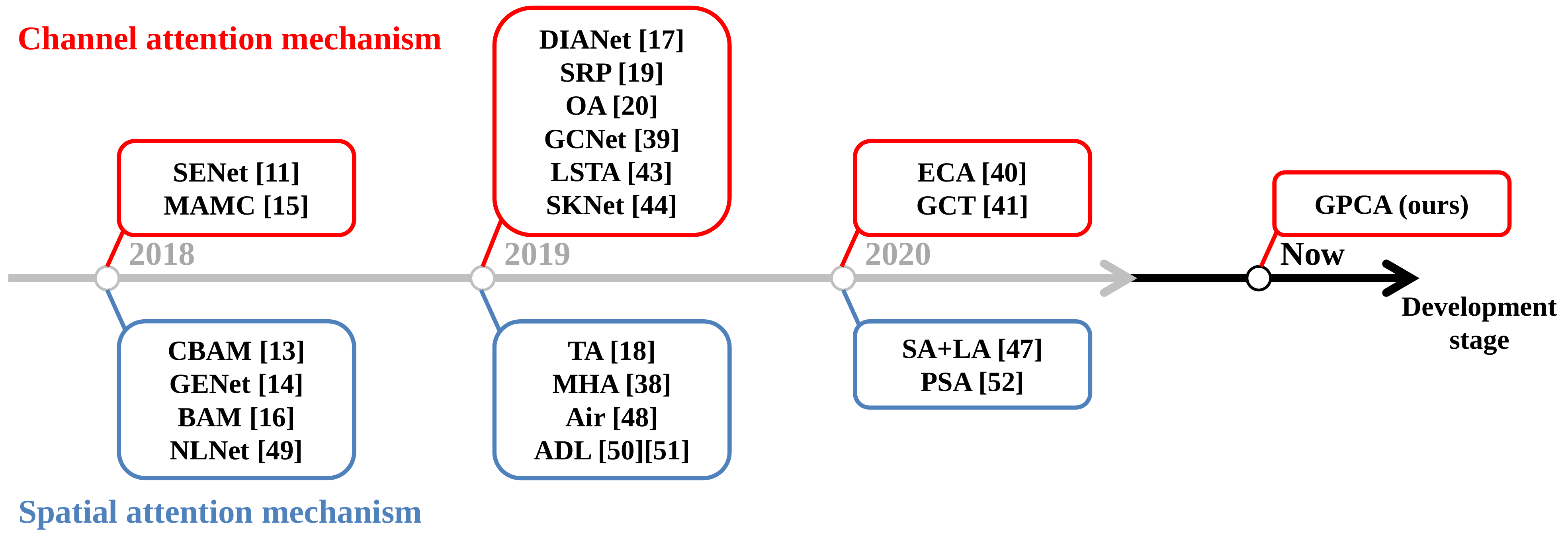}
  \vspace{-3mm}
  \caption{Development stages of attention mechanisms for visual tasks in recent years. The traditional attention mechanisms are commonly based on convolutional operations and pooling, while the GPCA module is derived within the probabilistic framework.}\label{fig:development}
  \vspace{-6mm}
\end{figure*}

Relevant work~\cite{hu2018squeeze,hu2018gather-excite,woo2018cbam,sun2018multi-attention,park2018bam,huang2019dianet,yang2019two-level,luo2019stochastic,lopez2019pay} have proposed several attention modules in different implementation ways. {However, the reasons why the conventional attention modules can learn the importance weights of the channels and/or regions are difficult to be explained, although the authors of those papers described their own understandings of their proposed modules, respectively. One way for promoting the explainability of attention mechanisms is to design an attention mechanism under the probabilistic framework, since it is well-known in its mechanism's intuitiveness and reasonableness~\cite{bishop06}.}

In this paper, we propose a novel attention mechanism, namely~\emph{Gaussian process embedded channel attention (GPCA)}, to clarify the channel attention mechanism intuitively and reasonably in a probabilistic way. The graphical model of the GPCA is shown in Figure \ref{fig:graphicalmodel}. Generally speaking, the channel attention masks are bounded in the interval $[0,1]$, which is a common boundary for traditional attentions. Hence, it is intuitive that we assume these attention mask values are generated from beta distributions. {However, the beta distribution cannot be directly integrated into the end-to-end training of the CNNs with an analytically tractable (closed-form) solution. The reason is that we need to find a differentiable function in terms of the distribution parameters for backpropagation. When we apply the beta variable in the CNN training, such differentiable function is infeasible due to the gamma function (defined by integration of the variable) in beta probability density function (PDF).} To tackle the above difficulty, we utilize an appropriate approximation of the beta distribution to solve this problem. We name this approximation as ``Sigmoid-Gaussian approximation'', in which the Gaussian distributed variables can be transferred and bounded into the interval $[0,1]$ by a Sigmoid function. Given that a channel attention module learns the weights of channels by computing their correlations precisely within end-to-end training, Gaussian process (GP)~\cite{bishop06,Rasmussen2006gaussian}, which is a popular Bayesian learning framework, is integrated into the GPCA module as the prior of the Gaussian distributed variables. This aims to learn the correlations among different channels in an explicit way. In this case, the informative channels can be emphasized by the large values from the attention masks, extracted by the GPCA module. 

Experimental results show that the proposed GPCA module can achieve the state-of-the-art performance on five datasets for the task of image classification. Meanwhile, we further evaluate the GPCA module in weakly supervised object localization (WSOL), object detection, {and semantic segmentation tasks} and demonstrate its reliable state-of-the-art performance.

\vspace{-4mm}
\section{Related Work}\label{sec:relatedwork}

Attention mechanisms are well-known and play important roles in various visual tasks including image retrieval~\cite{chen2019hybrid,fang2019bilinear}, object detection~\cite{zhao2019pyramid,chen2018reverse,linsley2018global}, person re-identification~\cite{chen2019mixed,chen2019self,yan2018multi-level}, visual tracking~\cite{zhu2018end-to-end}, image super-resolution~\cite{muqeet2019hybrid,jang2019densenet,dai2019second}, semantic segmentation~\cite{fu2019dual,huang2019ccnet,lu2019see}, and especially in image classification task~\cite{hu2018squeeze,hu2018gather-excite,woo2018cbam,sun2018multi-attention,park2018bam,huang2019dianet,yang2019two-level,luo2019stochastic,lopez2019pay,bello2019attention}. Most of the aforementioned works discuss what is the most effective way to introduce the attention mechanisms. These researches mainly concentrate on forcing CNNs to focus on the informative channels and/or the regions with convolutional feature maps. In general, the aforementioned work can be divided into three categories,~\emph{i.e.}, channel attention mechanisms, spatial attention mechanisms, and channel-spatial attention mechanisms. Usually, the spatial attention is combined with the channel attention in the CNN frameworks to further regularize the feature maps. Hence, we discuss the spatial and the channel-spatial attention mechanisms together in Section~\ref{ssec:sam}. We show the development stages of the attention mechanisms in Figure~\ref{fig:development}.

\vspace{-4mm}
\subsection{Channel Attention Mechanism}\label{ssec:cam}

Squeeze-and-excitation network (SENet)~\cite{hu2018squeeze}, which has been widely used in various network architectures, was proposed to investigate the channel relationship by a $2$-layer fully-connected (FC) network, yielding significant improvements in the image classification task. By proposing the squeeze-and-excitation (SE) mechanism, the SENet squeezes the feature maps across their spatial dimensions to introduce a channel descriptor and excites the embedded channel descriptor to per-channel modulation weights by a self-gating mechanism. The work in~\cite{sun2018multi-attention} was applied in CNNs, consisting of one-squeeze multi-excitation (OSME) modules and a multi-attention multi-class constraint (MAMC) for fine-grained image classification. The OSME modules are the extensions of the architecture of the SENet. They perform as channel attention modules for multiple discriminative regions discovered in the input images and improve the overall recognition performance. In addition, stochastic region pooling (SRP)~\cite{luo2019stochastic}, a channel attention module, randomly selected the square regions from the feature maps to extract the channel descriptors and forced them more representative and diverse. Then, it excited the channel descriptors in a similar way as the SENet to produce channel attention masks. Cao~\emph{et al.}~\cite{cao2019gcnet} replaced the global average pooling (GAP) of the SENet by a context modeling to describe the global context and named it as the global context network (GCNet). Wang~\emph{et al.}~\cite{wang2020ecanet} proposed a local cross-channel interaction strategy for channel attention mechanism to introduce a lightweight SENet and named it as efficient channel attention (ECA) module. Gated channel transformation (GCT)~\cite{yang2020gated} is another variant of the SENet, which combines a global context embedding, a channel normalization, and a gating adaptation for efficient and accurate contextual information modeling.

There are still other attention-related works which import the channel attention mechanisms with special motivations. Dense-and-implicit attention network (DIANet)~\cite{huang2019dianet} introduced a modified long short term memory (LSTM)~\cite{hochreiter1997long} model as the channel attention module for communication of the information among different convolutional layers and enhanced the generalization capacity of CNNs by repeatedly fixing the information. Sudhakaran~\emph{et al.}~\cite{sudhakaran2019lsta} directly integrated the attention mechanism into the LSTM for egocentric action recognition and named it as the long short-term attention (LSTA). The work in~\cite{lopez2019pay} captured the channel attentions after each convolutional block for rectifying the output likelihood distributions of the whole CNN, rather than the corresponding feature maps in the blocks. This can be considered as an output attention (OA) mechanism. In addition, the channel attention mechanism has also been utilized for convolutional kernel selection. Selective kernel network (SKNet)~\cite{li2019selective} was proposed to dynamically adjust the receptive field sizes of convolutional layers in CNNs by weighting the parallel convolutional kernels with different kernel sizes in a convolutional layer. Chen at al.~\cite{chen2020dynamic} proposed a dynamic convolution with the SENet module generating the weights of the convolution kernels.

\vspace{-4mm}
\subsection{Spatial Attention Mechanism}\label{ssec:sam}

Bottleneck attention module (BAM)~\cite{park2018bam} created two separate and parallel pathways,~\emph{i.e.}, the channel-wise and the spatial-wise ones, to construct the channel and spatial attentions. It combined them together as an element-wise attention. Meanwhile, different from the BAM, the convolutional block attention module (CBAM)~\cite{woo2018cbam} inferred the element-wise attention maps by cascade-connected channels and spatial attentions for the adaptive refinement of the intermediate feature maps. Gather-excite network (GENet)~\cite{hu2018gather-excite} introduced a parametric gather-excite operator pair into the residual blocks in ResNet~\cite{he2016deep}, which is considered as a generalization of the aforementioned SENet. It yielded further performance gains. The gathering operation aggregates the feature responses across the spatial neighbourhoods in a feature map, which is a two-dimensional pooling. The exciting operation redistributes the pooled information to local features via nearest neighbour interpolation. In addition, two-level attention (TA) module~\cite{yang2019two-level} implemented the channel and spatial attentions which characterized the object-level and the pixel-level attention, respectively. It combined these two attentions through a second-order response transform called bilinear pooling~\cite{lin2015bilinear} for fine-grained visual recognition. Zheng~\emph{et al.}~\cite{zheng2020blobal} proposed a global and local knowledge-aware attention network, combining a statistic-based attention (SA) and a learning-based attention (LA) together, for action recognition. Yang~\emph{et al.}~\cite{yang2019attention} introduced an attention inspiring receptive-fields (Air) module in a encoder-decoder architecture and implemented it by convolutional layers and a bilinear pooling.

Self-attention~\cite{vaswani2017attention} is a special (channel-)spatial attention mechanism, which yields the weighted averaged values computed from the feature maps themselves. Non-local neural network (NLNet)~\cite{wang2018non} combined a trilinear pooling with $1\times1$-convolutional layers and softmax functions to capture the long-range dependencies with deep neural networks in videos. Attention-based dropout layer (ADL)~\cite{cheo2019attention,choe2020attention} applied the self-attention mechanism by a channel-wise pooling and generated the spatial attention maps for weakly supervised object localization. Bello~\emph{et al.}~\cite{bello2019attention} developed a two-dimensional relative self-attention mechanism which introduces multihead-attention (MHA) as a key component. Zhao et al.~\cite{zhao2020exploring} explored variations of self-attention for image recognition and proposed a pairwise self-attention (PSA) network to improve the robustness and the generalization ability of the attention mechanism.

\vspace{-4mm}
\section{Gaussian Process Embedded Channel Attention (GPCA)}\label{sec:gpca}

\begin{figure}[!t]
  \centering
  \includegraphics[width=0.95\linewidth]{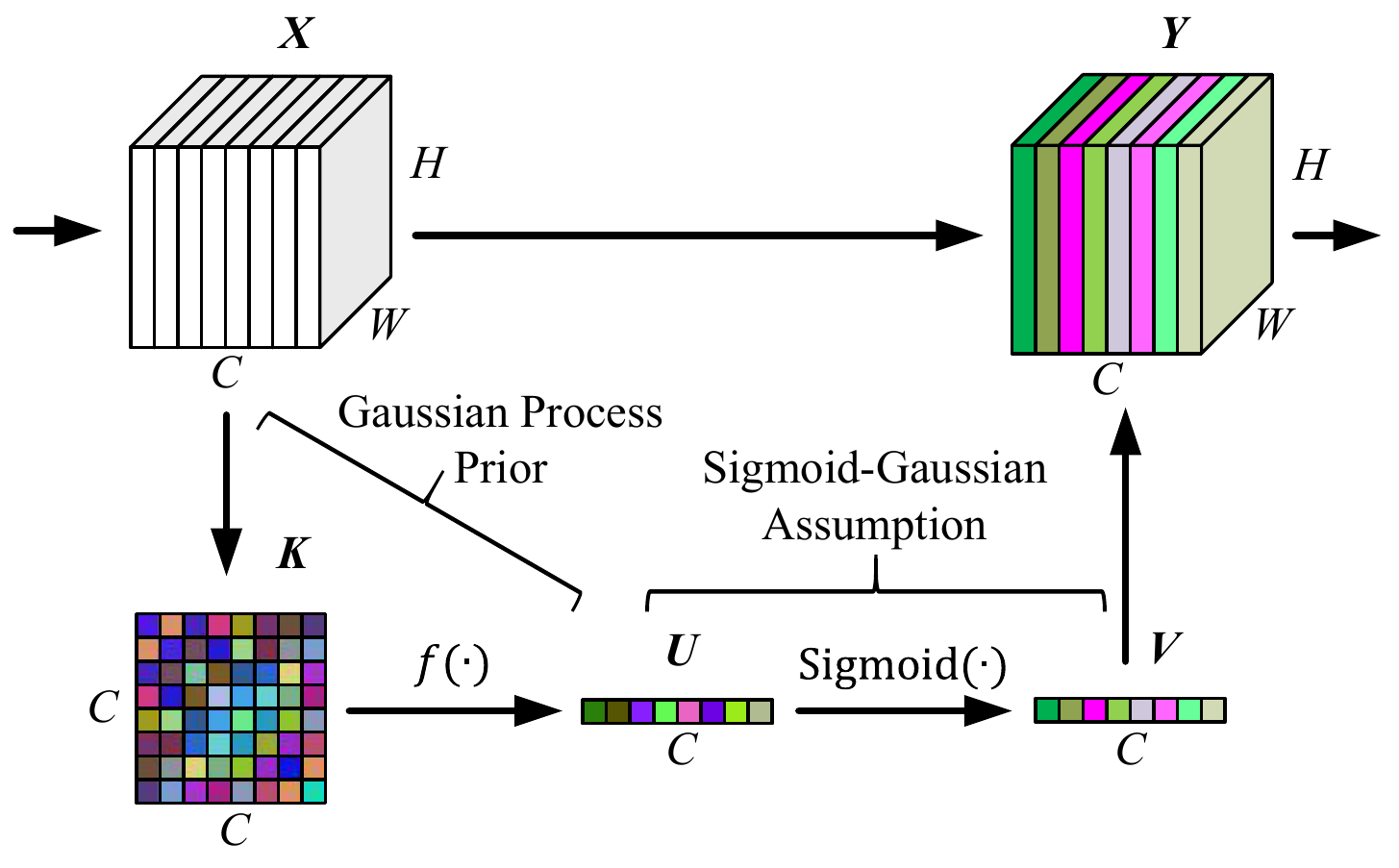}
  \vspace{-4mm}
  \caption{Structure of the proposed GPCA module.}\label{fig:structure}
  \vspace{-6mm}
\end{figure}

In this section, we introduce the Gaussian process embedded channel attention (GPCA) module, for the purpose of modeling the channel attention mechanism in a probabilistic way and interpreting it intuitively and reasonably.

\begin{figure*}[!t]
`\vspace{-2mm}
  \centering
  \includegraphics[width=0.99\linewidth]{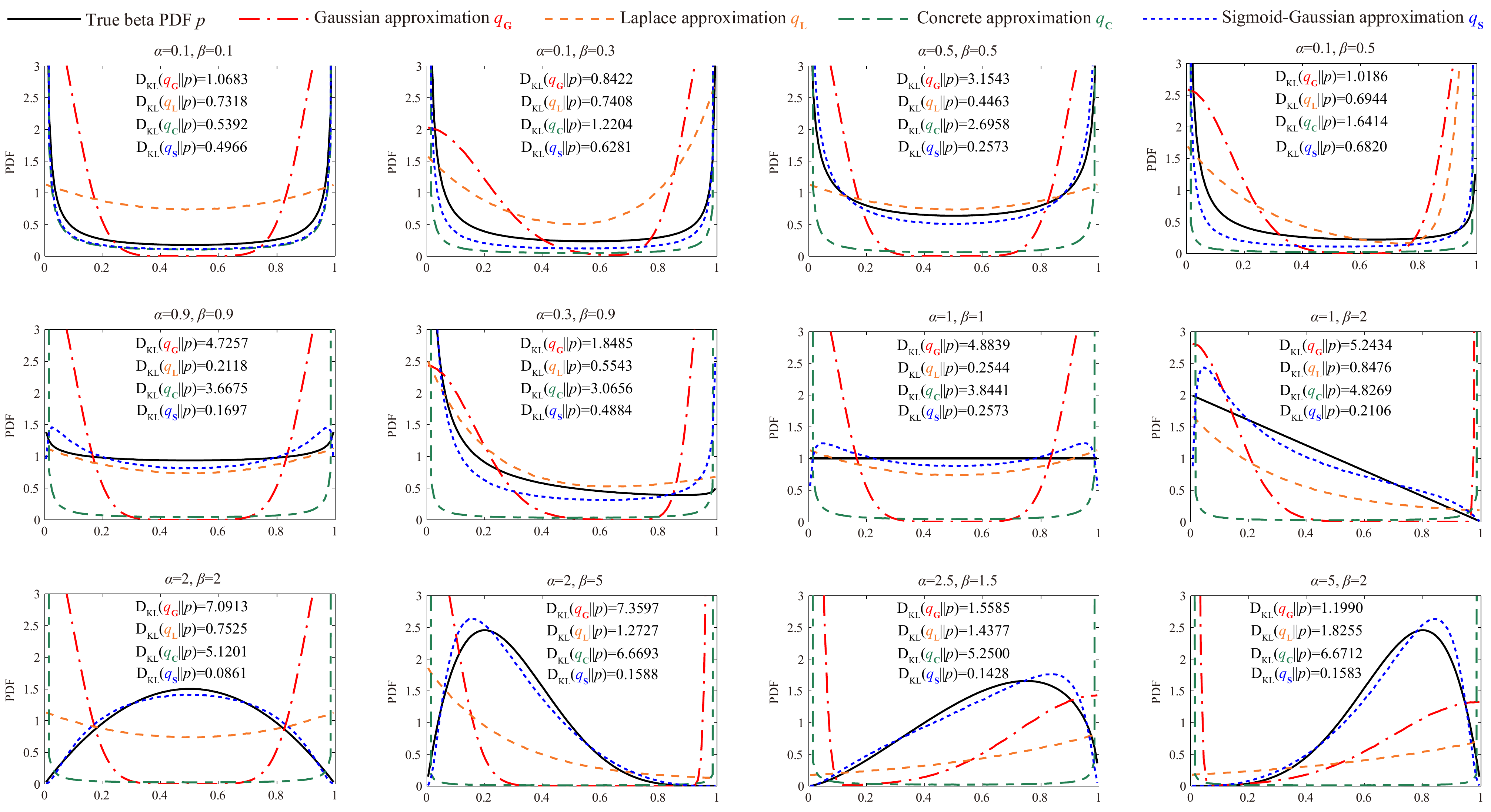}
  \caption{Probability density function (PDF) comparisons of the proposed Sigmoid-Gaussian approximation ($q_S$) with the beta distribution ($p$), the Gaussian approximation ($q_G$) and the Laplace approximation ($q_L$) used in~\cite{xie2019soft}, {and concrete approximation ($q_C$) proposed in~\cite{maddison2017the}}. The Kullback-Leibler (KL) divergences are calculate from all the approximations to their corresponding true beta PDFs with different $\alpha$ and $\beta$.}
  \label{fig:betaapprox}
  \vspace{-4mm}
\end{figure*}

A GPCA module can learn the correlations between channels in CNNs via a Gaussian process~\cite{bishop06} and calibrate the input feature maps by multiple channel-wise masks. Formally, we define the input feature maps by $\boldsymbol{X}\in R^{C\times W\times H}$ as shown in Figure~\ref{fig:structure}, where $C$, $W$, and $H$ are the number of channels, the width, and the height of the feature maps, respectively. Meanwhile, the final output $\boldsymbol{Y}\in R^{C\times W\times H}$ of the GPCA module is obtained by scaling $\boldsymbol{X}$ with the attention mask vector $\boldsymbol{V}=[v_1,\cdots,v_C]^{\text{T}}$ as
\begin{footnotesize}
\begin{equation}\label{eq:y}
  \boldsymbol{y}_c=v_c\cdot\boldsymbol{x}_c,\ c=1,\cdots,C,
\end{equation}\vspace{-4mm}
\end{footnotesize}

\noindent where $\boldsymbol{x}_c$ and $\boldsymbol{y}_c$ are the channels of $\boldsymbol{X}$ and $\boldsymbol{Y}$, respectively.

\vspace{-4mm}
\subsection{Attention Modeling via Sigmoid-Gaussian Approximation}\label{ssec:betaapprox}

The element $v_c$ of the attention mask vector $\boldsymbol{V}$ is commonly set as a $[0,1]$-bounded variable~\cite{hu2018squeeze,hu2018gather-excite,woo2018cbam,park2018bam} to scale the input feature maps with the normalized importance weights. Thus, we assume that the elements in $\boldsymbol{V}$ follow beta distributions to represent the importance of the channels. The probability density function (PDF) of the beta distribution is defined as
\begin{footnotesize}
\begin{equation}\label{eq:betadistrib}
  \text{Beta}(x;\alpha,\beta)=\frac{x^{\alpha-1}(1-x)^{\beta-1}}{\text{B}(\alpha,\beta)},
\end{equation}\vspace{-4mm}
\end{footnotesize}

\noindent where $\text{B}(\alpha,\beta)=\frac{\Gamma(\alpha)\Gamma(\beta)}{\Gamma(\alpha+\beta)}$ and $\Gamma(\cdot)$ is the Gamma function. The beta distribution, as a member of the exponential family, is defined in a bounded interval,~\emph{i.e.}, $[0,1]$, which is consistent with the definition of the channel attention. Meanwhile, the beta distribution is used for describing the statistical behavior of the percentages or the proportions in Bayesian learning such as certainty or importance. This further satisfies the principles of the attention modeling. In addition, it has flexible shapes (\emph{i.e.}, bell, ``U'', or uniform shapes) depending on different values of the parameters $\alpha$ and $\beta$, which makes it flexible for the channel attention learning in general.

However, the beta distribution cannot be directly applied in the CNN training~\cite{xie2019soft}. This is because when we apply the beta variable in the CNN training (normally using gradient descent-based algorithms), it is required to find a differentiable function $g(\cdot)$ for the backpropagation. With the beta PDF in~\eqref{eq:betadistrib}, we have to compute the derivatives of the logarithm of the Gamma functions during the backpropagation. As the Gamma function $\Gamma(\cdot)$ is defined by
\begin{footnotesize}
\begin{equation}
  \Gamma(x)=\int_0^{\infty}z^{x-1}e^{-z}dz,
\end{equation}\vspace{-4mm}
\end{footnotesize}

\noindent where the derivative of the logarithm of the Gamma function (\emph{i.e.}, the digamma function) is~\emph{again} defined by an integration form according to the Leibniz integral rule. Since $g(\cdot)$ contains the derivative of the logarithm of the Gamma function, it is not feasible to find a closed-form function to calculate $g(\cdot)$ explicitly. Although we can apply some numerical sampling methods,~\emph{e.g.}, Markov chain Monte Carlo (MCMC), to numerically simulate $g(\cdot)$, the computational cost will be high, especially when dealing with high-dimensional data. In other words, we cannot directly employ the beta distribution in the CNN training, as the closed-form solution for $g(\cdot)$ is unavailable. Alternatively, we introduce an approximation $q(\alpha,\beta)$, named as Sigmoid-Gaussian approximation, for the beta distribution to solve the unfeasible issue in backpropagation. A random variable $v\sim q(\alpha,\beta)$ following the distribution assumption can be generated by transferring a Gaussian distributed variable $u\sim\mathcal{N}(\mu,\sigma^2)$ as
\begin{footnotesize}
\begin{equation}\label{eq:ugeneratesv}
  v=\text{Sigmoid}(u)=\frac{1}{1+e^{-u}},
\end{equation}\vspace{-4mm}
\end{footnotesize}

\noindent and {$u=\ln v-\ln(1-v)$. In this case, parameters $\alpha$ and $\beta$ can be obtained by matching the first- and second-order moments between $u$ and $v$ as}
\begin{footnotesize}
\begin{equation}\label{eq:mm}
  \begin{cases}
    \mu=\phi(\alpha)-\phi(\beta)\\
    \sigma^2=\phi_1(\alpha)+\phi_1(\beta)
  \end{cases},
\end{equation}\vspace{-4mm}
\end{footnotesize}

\noindent where $\phi(\cdot)$ and $\phi_1(\cdot)$ are the digamma and trigamma functions, respectively.\footnote{Please refer to \url{https://en.wikipedia.org/wiki/Beta_distribution}.} {For better understanding, we derive the equations in~\eqref{eq:mm} by} 
\begin{footnotesize}
\begin{align}
    \mu&=\text{E}[u]=\text{E}[\ln v-\ln(1-v)]=\text{E}[\ln v]-\text{E}[\ln(1-v)]\nonumber\\
    &=(\phi(\alpha)-\phi(\alpha+\beta))-(\phi(\beta)-\phi(\alpha+\beta))=\phi(\alpha)-\phi(\beta),\label{eq:mumm}\\
    \sigma^2&=\text{Var}[u]=\text{Var}[\ln v-\ln(1-v)]=\text{Var}\left[\ln\left( \frac{v}{1-v}\right)\right]\nonumber\\
    &=-\text{Cov}\left[\ln\left(\frac{v}{1-v}\right),\ln\left(\frac{1-v}{v}\right)\right]=\phi_1(\alpha)+\phi_1(\beta),\label{eq:sigmamm}
\end{align}\vspace{-4mm}
\end{footnotesize}

\noindent {with the properties of the beta distribution.} Since~\eqref{eq:mumm} and~\eqref{eq:sigmamm} illustrate a one-to-one mapping between ($\alpha$, $\beta$) and ($\mu$, $\sigma$), we can optimise ($\mu$, $\sigma$) instead of ($\alpha$, $\beta$) to facilitate backpropagation training.

For $v$ in \eqref{eq:ugeneratesv}, with the principle of integration by substitution, we have
\begin{footnotesize}
\begin{align}\label{eq:proofnormalized}
    \int p(u)du\nonumber   =&\int\frac{1}{\sqrt{2\pi}\sigma}e^{-\frac{\left(\ln v-\ln(1-v)-\mu\right)^2}{2\sigma^2}} \cdot\left(\frac{1}{v(1-v)}\right)dv\nonumber\\
    =&\int p(v)dv\nonumber\\
    =&1.
\end{align}
\vspace{-6mm}
\end{footnotesize}

To demonstrate the effectiveness of the Sigmoid-Gaussian approximation, we conducted several groups of simulations with different parameters, as shown in Figure~\ref{fig:betaapprox}. We compare the PDF $q_S$ of the Sigmoid-Gaussian approximation with {three} different approximation methods ({including Gaussian approximation $q_G$ and Laplace approximation $q_L$ used in~\cite{xie2019soft}, and concrete approximation $q_C$~\cite{maddison2017the}}) and the true beta PDFs $p$ with parameters $\alpha$ and $\beta$. {The beta approximation methods can be divided into two groups, \emph{i.e.}, single distributions (including Gaussian approximation and Laplace approximation) and distributions via variable transfer (including concrete approximation and the proposed Sigmoid-Gaussian approximation). The proposed Sigmoid-Gaussian approximation can approach the shapes (\emph{e.g.}, bell, ``U'', skewed bell shapes) of the true beta distribution with different parameters and has the same interval (\emph{i.e.}, $v\in[0,1]$) with a beta distributed variable. This demonstrates the proposed Sigmoid-Gaussian approximation can be a beta approximation solution.} In addition, the Kullback-Leibler (KL) divergences from the approximated PDFs $q_S$, $q_G$, $q_L$, or $q_C$ to the true beta PDFs $p$ are calculated with parameters $\alpha$ and $\beta$ in each group. It can be observed that $q_S$ yields more accurate approximation to the corresponding $p$ with smaller KL divergences. This clearly demonstrates the effectiveness of the Sigmoid-Gaussian approximation.

\vspace{-3mm}
\subsection{Channel Correlation Modeling via GP Prior}\label{ssec:gpprior}
\vspace{-1mm}

{In order to adaptively model the correlations between channels, we introduce a Gaussian process (GP) prior $\mathcal{GP}(\cdot)$ with the input feature map $\boldsymbol{X}$ for $\boldsymbol{U}$ as $\boldsymbol{U}\sim\mathcal{GP}(\boldsymbol{X})$. Here we treat the channels in $\boldsymbol{X}$ as the samples in $\mathcal{GP}(\cdot)$. According to the definition of the Sigmoid-Gaussian assumption for beta approximation in Section~\ref{ssec:betaapprox}, for the $c^{th}$ channel, the parameters $\alpha_c$ and $\beta_c$ of the beta distribution can be matched by $\mu_c$ and $\sigma_c^2$ in~\eqref{eq:mm} and we can treat the Gaussian processing as the prior of the parameters in the beta distribution.}

Although we cannot obtain any exact targets of attention masks in CNNs, we are able to learn the channel correlations in $\boldsymbol{X}$ by $\boldsymbol{a}_c\in R^{1\times (C-1)}$ for the $c^{th}$ channel to others, which is defined as
\begin{footnotesize}
\begin{equation}\label{eq:astar}
  \boldsymbol{a}_c=\boldsymbol{K}_{c,i\neq c}\left(\boldsymbol{K}_{i\neq c,i\neq c}+\delta^{-1}\boldsymbol{I}_{C-1}\right)^{-1},
\end{equation}\vspace{-4mm}
\end{footnotesize}

\noindent where $\delta$ represents the precision of the Gaussian noise in the Gaussian process. For the purpose of adapting non-informative prior, we set it as a large number (\emph{e.g.}, $1\times10^6$) in practice. $\boldsymbol{I}_{C-1}$ is an $(C-1)\times(C-1)$ identity matrix. $\boldsymbol{K}_{c,i\neq c}\in R^{1\times(C-1)}$ and $\boldsymbol{K}_{i\neq c,i\neq c}\in R^{(C-1)\times(C-1)}$ are parts of the Gram matrix $\boldsymbol{K}\in R^{C\times C}$ in the Gaussian process~\cite{bishop06,Rasmussen2006gaussian} with elements $K_{c,c'}, c,c'=1,\cdots,C$, defined as
\begin{footnotesize}
\begin{align}\label{eq:k}
  K_{c,c'}&=k(\boldsymbol{x}_c,\boldsymbol{x}_{c'})\nonumber\\
  &=\underbrace{\theta_0 e^{-\theta_1||\boldsymbol{x}_c-\boldsymbol{x}_{c'}||^2}\vphantom{\boldsymbol{x}^{\text{T}}_{c'}}}_{\text{Gaussian\,kernel}} +\underbrace{\theta_2\vphantom{\boldsymbol{x}^{\text{T}}_{c'}}}_{\text{Bias}}+\underbrace{\theta_3\boldsymbol{x}_c\boldsymbol{x}^{\text{T}}_{c'}}_{\text{Linear\,kernel}},
\end{align}\vspace{-4mm}
\end{footnotesize}

\noindent where $k(\boldsymbol{x}_c,\boldsymbol{x}_{c'})$ is a kernel function with nonnegative parameters $\boldsymbol{\Theta}=\left\{\theta_0,\theta_1,\theta_2,\theta_3\right\}$ and consists of a Gaussian kernel, a linear kernel, and a bias. In practice, we can either have $\boldsymbol{\Theta}$ fixed or set $\theta_i=e^{\tilde{\theta}_i}$ and optimize $\tilde{\theta}_i$ to satisfy the nonnegative constraints. In this paper, $\theta_i$ is implicitly optimized by $\tilde{\theta}_i$ in model training.

\noindent Then, we assume that the importance weight vector $\boldsymbol{U}=\left[u_1,\cdots,u_C\right]^{\text{T}}$ in Figure \ref{fig:structure} that the $c^{th}$ channel $u_c$,~$c=1,\cdots,C$, follows Gaussian distribution as
\begin{footnotesize}
\begin{equation}\label{eq:u}
  u_c\sim\mathcal{N}(A_c,B_c),
\end{equation}\vspace{-4mm}
\end{footnotesize}

\noindent where $A_c,B_c$ are the mean and the variance of $u_c$, respectively. According to~\cite{bishop06}, they can be calculated as
\begin{footnotesize}
\begin{align}
  A_c&=\frac{1}{C-1}\sum_{c'=1,c'\ne c}^C\tilde{a}_{c',c},\label{eq:ac}\\
  B_c&=K_{c,c}-\boldsymbol{a}_c\boldsymbol{K}^{\text{T}}_{c,i\neq c},\label{eq:bc}
\end{align}\vspace{-4mm}
\end{footnotesize}

\noindent where the row vector $\tilde{a}_{c}\in R^{1\times C}$ is obtained by inserting a constant before index $c$ of the row $a_c$ as
\begin{footnotesize}
\begin{equation}\label{eq:atildec}
  \tilde{a}_{c,i}=
  \begin{cases}
    a_{c,i} & 1\le i<c \\
    \epsilon  & i=c \\
    a_{c,i-1} & c<i\le C
  \end{cases},
\end{equation}\vspace{-4mm}
\end{footnotesize}

\noindent where $\epsilon$ is a constant. Note that the padding constant $\epsilon$ in~\eqref{eq:atildec} is for clearly clarifying the correspondence of the columns in each row, as $a_{c,i}$ corresponds to channel $i$ when $1\leq i<c$, $a_{c,i-1}$ corresponds to channel $i$ when $c<i\le C$, and no elements in $a_{c}$ correspond to channel $c$. In~\eqref{eq:ac}, $\tilde{a}_{c,c}$ is~\emph{not} involved in the calculation of $A_c$, which means that the paddings can use any value and we opt to use zeros here for the purpose of simplification in calculation.

{As $\boldsymbol{a}_{c}$ for each channel is calculated one-by-one in practice, we can obtain the correlations between any channel and the rest ones. Then, these correlations are averaged as the expected values of the channel attention weights, which is shown in~\eqref{eq:ac}.} With the aforementioned one-by-one strategy, the prior of each channel is set as a one-dimension Gaussian distribution. To specify, we fix $\boldsymbol{K}$ (during each iteration $\boldsymbol{K}$ is not updated when calculating $\boldsymbol{a}_{c}$ with different $c$) and consider each channel as the theoretical ``test input'' of the Gaussian process. Meanwhile, the other channels are considered as the ``training inputs'' of the Gaussian process.

In practice, the expectation of the random variable $v_c$ is considered as the final GPCA mask of the $c^{th}$ channel, which is calculated by~\cite{bishop06}
\begin{footnotesize}
\begin{equation}\label{eq:vmean}
  \text{E}\left[v_c\right]\approx\text{Sigmoid}\left(\frac{A_c}{\sqrt{1+\frac{\pi}{8}B_c}}\right).
\end{equation}\vspace{-4mm}
\end{footnotesize}

{The algorithm of the GPCA is summarized in Algorithm~\ref{alg:method}.}

\newlength{\textfloatseptemp}
\setlength{\textfloatseptemp}{\textfloatsep}
\setlength{\textfloatsep}{10pt}

\begin{algorithm}[!t]
    \caption{{Gaussian Process Embedded Channel Attention (GPCA)}}\label{alg:method}
    \begin{algorithmic}[1]
        \Require $\boldsymbol{X}$: input channels.
        \Ensure $\boldsymbol{Y}$: output channels.
        \State Calculate Gram matrix $\boldsymbol{K}$ in~\eqref{eq:k}.
        \For{$c = 1 \to C$}
            \State Calculate $\boldsymbol{a}_c$ in~\eqref{eq:astar}.
            \State Calculate $B_c$ in~\eqref{eq:bc}.
        \EndFor
        \For{$c = 1 \to C$}
            \State Calculate $A_c$ in~\eqref{eq:ac}.
        \EndFor
        \State Calculate $\boldsymbol{V}$ in~\eqref{eq:vmean}.
        \State Calculate $\boldsymbol{Y}$ in~\eqref{eq:y}.
    \end{algorithmic}
\end{algorithm}

\subsubsection{{Derivation of GP for Channel Correlation Modeling}}\label{sssec:gpderivation}

{We are motivated by the assumption that modeling the channel correlation and then calculating the channel attention masks is equivalent to a GP for regression with pseudo targets. Here, the pseudo targets are based on the aspiration that the channel attention mechanism can reflect the actual correlation of each channel and reconstruct the pseudo target of one channel according to those of the other channels and the corresponding correlations.}

{Specifically, taking the $c^{th}$ channel $\boldsymbol{x}_c$ as a test sample with its pseudo target $z_c$, we define the pseudo targets of the other $(C-1)$ channels as $\boldsymbol{z}_{i\neq c}$. {Here, we can transform $z_{c}$ to a vector $\boldsymbol{z}_{c}$ and set $\boldsymbol{z}_{c}=\boldsymbol{x}_{c}$, for the purpose of transforming the general regression task we defined into an autoregression one. In this case, reconstructing one channel with all the others can be considered as modeling the channel correlation and then calculating the channel attention masks. Since we can independently analyze each dimension of the vector $\boldsymbol{z}_{c}$, we show the derivation of the GP in terms of scalar $z_{c}$ in the following paragraphs for better demonstration.} We further design an aggregation function $\Phi(\boldsymbol{X}_{i\neq c})\in R^{M\times(C-1)}$ to map $\boldsymbol{X}_{i\neq c}$ into an $M$-dimensional feature space. Here, the relationship between the kernel function $K_{\cdot,\cdot}$ in~\eqref{eq:k} and the aggregation function $\Phi(\cdot)$ is as $K_{c,c^{'}}=k(\boldsymbol{x}_c,\boldsymbol{x}_{c^{'}})=\boldsymbol{\Phi}(\boldsymbol{x}_c)^{\text{T}}\boldsymbol{\Phi}(\boldsymbol{x}_{c^{'}})=\boldsymbol{\Phi}_c^{\text{T}}\boldsymbol{\Phi}_{c^{'}}$. Then, the relation between $\boldsymbol{z}_{i\neq c}\in R^{(C-1)\times1}$ and $\Phi(\boldsymbol{X}_{i\neq c})$ with parameters $\boldsymbol{w}_c\in R^{M\times1}$ is}
\begin{footnotesize}
\begin{equation}
    \boldsymbol{z}_{i\neq c}=\underbrace{\Phi(\boldsymbol{X}_{i\neq c})^{\text{T}}\boldsymbol{w}_c}_{f(\boldsymbol{X}_{i\neq c})}+\zeta\boldsymbol{1}_{C-1},
\end{equation}\vspace{-4mm}
\end{footnotesize}

\noindent{where $f(\cdot)$ is the function for regression, the elements in the parameter vector $\boldsymbol{w}_c$ follow independent standard normal distributions, the noise term $\zeta$ follows a Gaussian distribution with zero mean and $\delta$ precision, and $\boldsymbol{1}_{C-1}$ is a $(C-1)\times1$ vector that all the elements equal one. Then, $z_c$ is predicted by $f(\boldsymbol{x}_c)$ in the test phase as} 
\begin{footnotesize}
\begin{equation}
    z_c=\underbrace{\Phi(\boldsymbol{x}_c)^{\text{T}}\boldsymbol{w}_c}_{f(\boldsymbol{x}_c)}+\zeta.
\end{equation}\vspace{-4mm}
\end{footnotesize}

{To estimate $\boldsymbol{w}_c$, the likelihood is calculated as}
\begin{footnotesize}
\begin{equation}
    p(\boldsymbol{z}_{i\neq c}|\boldsymbol{X}_{i\neq c},\boldsymbol{w}_c)=\mathcal{N}(\boldsymbol{\Phi}^{\text{T}}\boldsymbol{w}_c,\delta^{-1}\boldsymbol{\boldsymbol{I}_{C-1}}),
\end{equation}\vspace{-4mm}
\end{footnotesize}

\noindent {where $\boldsymbol{\Phi}$ is the shorthand of $\Phi(\boldsymbol{X}_{i\neq c})$. According to the Bayes' rule, the posterior distribution $p(\boldsymbol{w}_c|\boldsymbol{X}_{i\neq c},\boldsymbol{z}_{i\neq c})$ is proportional to the product of the likelihood and the prior distribution of $\boldsymbol{w}_c$, which can be denoted as}
\begin{footnotesize}
\begin{align}
    &p(\boldsymbol{w}_c|\boldsymbol{X}_{i\neq c},\boldsymbol{z}_{i\neq c})\nonumber\\
    &\propto \exp{\left(-\frac{\delta}{2} (\boldsymbol{z}_{i\neq c}-\boldsymbol{\Phi}^{\text{T}}\boldsymbol{w}_c)^{\text{T}}(\boldsymbol{z}_{i\neq c}-\boldsymbol{\Phi}^{\text{T}}\boldsymbol{w}_c)\right)} \cdot \exp{\left(-\frac{1}{2}\boldsymbol{w}_c^{\text{T}}\boldsymbol{w}_c\right)}\nonumber\\
    &\propto \exp{\left(-\frac{1}{2}(\boldsymbol{w}_c-\bar{\boldsymbol{w}}_c)^{\text{T}}(\delta\boldsymbol{\Phi}\boldsymbol{\Phi}^{\text{T}}+\boldsymbol{I}_{M})(\boldsymbol{w}_c-\bar{\boldsymbol{w}}_c)\right)},
\end{align}\vspace{-4mm}
\end{footnotesize}

\noindent {where $\bar{\boldsymbol{w}}_c=\delta(\delta\boldsymbol{\Phi}\boldsymbol{\Phi}^{\text{T}}+\boldsymbol{I}_{M})^{-1}\boldsymbol{\Phi}\cdot\boldsymbol{z}_{i\neq c}$. In this case, the posterior distribution $p(\boldsymbol{w}_c|\boldsymbol{X}_{i\neq c},\boldsymbol{z}_{i\neq c})$ is}
\begin{footnotesize}
\begin{align}
    &p(\boldsymbol{w}_c|\boldsymbol{X}_{i\neq c},\boldsymbol{z}_{i\neq c})\nonumber\\
    &=\mathcal{N}\left(\delta(\delta\boldsymbol{\Phi}\boldsymbol{\Phi}^{\text{T}}+\boldsymbol{I}_{M})^{-1}\boldsymbol{\Phi}\cdot\boldsymbol{z}_{i\neq c},(\delta\boldsymbol{\Phi}\boldsymbol{\Phi}^{\text{T}}+\boldsymbol{I}_{M})^{-1}\right).
\end{align}\vspace{-4mm}
\end{footnotesize}

\noindent {Then, we can predict the pseudo target $\boldsymbol{z}_{c}$ of the $c^{th}$ channel by $\boldsymbol{x}_{c}$ according to $\{\boldsymbol{x}_i,z_i\}_{i\neq c}$ as}
\begin{footnotesize}
\begin{align}\label{eq:prediction}
    f(\boldsymbol{x}_{c})&=\boldsymbol{\Phi}_c^{\text{T}}\boldsymbol{w}_c\nonumber\\
    &\sim\mathcal{N}(\underbrace{\boldsymbol{\Phi}_c^{\text{T}}\delta(\delta\boldsymbol{\Phi}\boldsymbol{\Phi}^{\text{T}}+\boldsymbol{I}_{M})^{-1}\boldsymbol{\Phi}}_{\boldsymbol{a}_c}\cdot\boldsymbol{z}_{i\neq c},\underbrace{\boldsymbol{\Phi}_c^{\text{T}}(\delta\boldsymbol{\Phi}\boldsymbol{\Phi}^{\text{T}}+\boldsymbol{I}_{M})^{-1}\boldsymbol{\Phi}_c}_{B_c}),
\end{align}\vspace{-4mm}
\end{footnotesize}

\noindent {where $\boldsymbol{\Phi}_c$ is the shorthand of $\Phi(\boldsymbol{x}_{c})\in R^{M\times1}$.} {Here, $\boldsymbol{a}_c$ can model the relation between channels, since we obtain the mean of the (pseudo) prediction of the $c^{th}$ channel by weighted summation $\boldsymbol{z}_{i\neq c}$ by weights $\boldsymbol{a}_c$. Meanwhile, $B_c$ can be considered as the variance of the channel attentions, when transforming $z_{c}$ to a vector $\boldsymbol{z}_{c}$ and setting $\boldsymbol{z}_{c}=\boldsymbol{x}_{c}$. In other words, we explicitly model the estimator $\text{E}[\boldsymbol{x}_c|\boldsymbol{X}_{i\neq c}]$ that reconstructures $\boldsymbol{x}_c$ by $\boldsymbol{X}_{i\neq c}$ to compute the optimal attention values (\emph{i.e.}, importance weights) by following the certain criterion of the GP.}

{Here, we will derive the expressions of $\boldsymbol{a}_c$ and $B_c$ in~\eqref{eq:prediction}, respectively, as those given by~\eqref{eq:astar} and~\eqref{eq:bc}. Denote $\boldsymbol{\kappa}=\delta\boldsymbol{\Phi}\boldsymbol{\Phi}^{\text{T}}+\boldsymbol{I}_M$, we have}
\begin{footnotesize}
\begin{align}
   \boldsymbol{\kappa}\boldsymbol{\Phi}=&\left(\delta\boldsymbol{\Phi}\boldsymbol{\Phi}^{\text{T}}+\boldsymbol{I}_M\right)\boldsymbol{\Phi}=\delta\boldsymbol{\Phi}\underbrace{\left(\boldsymbol{\Phi}^{\text{T}}\boldsymbol{\Phi}+\delta^{-1}\boldsymbol{I}_{C-1}\right)}_{\boldsymbol{\tau}}.
\end{align}\vspace{-4mm}
\end{footnotesize}

\noindent {Following the principals of matrix algebra, it can be obtained that}
\begin{footnotesize}
\begin{align}
\label{Eq:Derivation of ac}
    \boldsymbol{\kappa}^{-1}\boldsymbol{\kappa}\boldsymbol{\Phi}\boldsymbol{\tau}^{-1}&=\boldsymbol{\kappa}^{-1}\delta\boldsymbol{\Phi}\boldsymbol{\tau}\boldsymbol{\tau}^{-1}\Longrightarrow \boldsymbol{\Phi}\boldsymbol{\tau}^{-1}=\boldsymbol{\kappa}^{-1}\delta\boldsymbol{\Phi}.
\end{align}\vspace{-4mm}
\end{footnotesize}

\noindent {With~\eqref{Eq:Derivation of ac}, $\boldsymbol{a}_c$ can be reformulated as}
\begin{footnotesize}
\begin{align}
    \boldsymbol{a}_c&=\boldsymbol{\Phi}_c^{\text{T}}\left(\boldsymbol{\kappa}^{-1}\delta\boldsymbol{\Phi}\right)=\boldsymbol{\Phi}_c^{\text{T}}\left(\boldsymbol{\Phi}\boldsymbol{\tau}^{-1}\right)\nonumber\\
    &=\underbrace{\boldsymbol{\Phi}_c^{\text{T}}\boldsymbol{\Phi}}_{\boldsymbol{K}_{c,i\neq c}}(\underbrace{\boldsymbol{\Phi}^{\text{T}}\boldsymbol{\Phi}}_{\boldsymbol{K}_{i\neq c,i\neq c}}+\delta^{-1}\boldsymbol{I}_{C-1})^{-1}.
\end{align}\vspace{-4mm}
\end{footnotesize}

{For the extension of $B_c$, we introduce the matrix inversion lemma (or called Woodbury, Sherman \& Morrison formula) as}
\begin{footnotesize}
\begin{align}
    &(\boldsymbol{Z}+\boldsymbol{U}\boldsymbol{W}\boldsymbol{V}^{\text{T}})^{-1}\nonumber\\
    &=\boldsymbol{Z}^{-1}-\boldsymbol{Z}^{-1}\boldsymbol{U}(\boldsymbol{W}^{-1}+\boldsymbol{V}^{\text{T}}\boldsymbol{Z}^{-1}\boldsymbol{U})^{-1}\boldsymbol{V}^{\text{T}}\boldsymbol{Z}^{-1},
\end{align}\vspace{-4mm}
\end{footnotesize}

\noindent {where $\boldsymbol{Z}$, $\boldsymbol{U}$, $\boldsymbol{W}$, and $\boldsymbol{V}$ are matrices. Here, we set $\boldsymbol{Z}=\boldsymbol{I}_{M}$, $\boldsymbol{U}=\boldsymbol{V}=\boldsymbol{\Phi}$, and $\boldsymbol{W}=\delta\boldsymbol{I}_{C-1}$, respectively and obtain}
\begin{footnotesize}
\begin{align}
    B_c&=\boldsymbol{\Phi}_c^{\text{T}}(\boldsymbol{\Phi}(\delta\boldsymbol{I}_{C-1})\boldsymbol{\Phi}^{\text{T}}+\boldsymbol{I}_{M})^{-1}\boldsymbol{\Phi}_c\nonumber\\
    &=\boldsymbol{\Phi}_c^{\text{T}}(\boldsymbol{I}_{M}-\boldsymbol{I}_{M}\boldsymbol{\Phi}(\delta^{-1}\boldsymbol{I}_{C-1}+\boldsymbol{\Phi}^{\text{T}}\boldsymbol{I}_{M}\boldsymbol{\Phi})^{-1}\boldsymbol{\Phi}^{\text{T}}\boldsymbol{I}_{M})\boldsymbol{\Phi}_c\nonumber\\
    &=\underbrace{\boldsymbol{\Phi}_c^{\text{T}}\boldsymbol{\Phi}_c}_{K_{c,c}}-\underbrace{\boldsymbol{\Phi}_c^{\text{T}}\boldsymbol{\Phi}(\boldsymbol{\Phi}^{\text{T}}\boldsymbol{\Phi}+\delta^{-1}\boldsymbol{I}_{C-1})^{-1}}_{\boldsymbol{a}_c}\underbrace{\boldsymbol{\Phi}^{\text{T}}\boldsymbol{\Phi}_c}_{\boldsymbol{K}_{c,i\neq c}^{\text{T}}}.
\end{align}
\end{footnotesize}

\setlength{\textfloatsep}{\textfloatseptemp}

\vspace{-6mm}
\subsubsection{{Derivation of Approximated Expectation of Sigmoid-Gaussian Approximation}} \label{sssec:expectationsigmoidgaussianderivation}

{The expectation of the variable $v_c$ can be calculated by}
\begin{footnotesize}
\begin{align}\label{eq:expectationsigmoid}
  \text{E}\left[v_c\right]&=\int_0^1 v_c q(v_c)d\,v_c\nonumber\\
  &=\int_{-\infty}^{+\infty}\text{Sigmoid}(u_c)\mathcal{N}(u_c;A_c,B_c)d\,u_c\nonumber\\
  &\approx\int_{-\infty}^{+\infty}\varphi(\sqrt{\frac{\pi}{8}}u_c)\mathcal{N}(u_c;A_c,B_c)d\,u_c\nonumber\\
  &=\varphi\left(\frac{A_c}{\sqrt{\frac{8}{\pi}+B_c}}\right)\approx\text{Sigmoid}\left(\frac{A_c}{\sqrt{1+\frac{\pi}{8}B_c}}\right),
\end{align}
\end{footnotesize}

\noindent {where}
\begin{footnotesize}
\begin{equation}\label{eq:cdfapprox}
  \varphi(x)=\int_{-\infty}^{x}\mathcal{N}(t;0,1)dt \approx\text{Sigmoid}\left(\lambda x\right)
\end{equation}\vspace{-4mm}
\end{footnotesize}

\noindent {is the cumulative distribution function of a standard normal distribution (or called probit function) with equality condition as $x=0$~\cite{mackey1992the}. The weight $\lambda=\sqrt{\frac{8}{\pi}}$ can be obtained by matching the first-order derivatives between $\varphi(x)$ and $\text{Sigmoid}\left(\lambda x\right)$ at $x=0$.} {Then the approximated expectation in~\eqref{eq:cdfapprox} is used for the final GPCA mask to the $c^{th}$ channel.}

\begin{table}[!t]
  \centering
  \caption{Statistics of image classification datasets including the class number, the training and test sample number, and the input image size.}
    \begin{tabular}{lcccc}
    \toprule
    Dataset & \#class & \#training & \#test & Image size \\
    \midrule
    Cifar-$10$ & $10$    & $50,000$   & $10,000$   & $32\times32$ \\
    Cifar-$100$ & $100$   & $50,000$   & $10,000$   & $32\times32$ \\
    \emph{mini}ImageNet & $100$   & $50,000$   & $10,000$   & $224\times224$ \\
    ImageNet-$32\times32$ & $1000$  & $1,281,123$  & $50,000$    & $32\times32$ \\
     ImageNet &  $1000$  &  $1,281,123$  &  $50,000$    &  $224\times224$ \\
    \bottomrule
    \end{tabular}
  \label{tab:clsdatasets}
  \vspace{-4mm}
\end{table}

\vspace{-4mm}
\section{Experiments on Image classification}\label{sec:classification}

In this section, we compare the proposed GPCA with {eight} recently proposed attention modules,~\emph{i.e.}, SENet~\cite{hu2018squeeze}, {NLNet~\cite{wang2018non}}, BAM~\cite{park2018bam}, CBAM~\cite{woo2018cbam}, SRP~\cite{luo2019stochastic}, OA~\cite{lopez2019pay}, ECA~\cite{wang2020ecanet}, and GCT~\cite{yang2020gated}, on five datasets for image classification.

\vspace{-4mm}
\subsection{Datasets}\label{ssec:datasets}

The five image datasets includes Cifar-$10$/-$100$~\cite{krizhevsky09cifar}, \emph{mini}ImageNet~\cite{vinyals2016matching}, ImageNet-$32\!\times\!32$~\cite{denoord2016pixel}, {ImageNet~\cite{russakovsky2015imagenet}} datasets. The summary of the datasets is listed in Table~\ref{tab:clsdatasets}. In the~\emph{mini}ImageNet dataset, $500$ and $100$ images of per class are respectively randomly selected from the full ImageNet dataset for training and test. The ImageNet-$32\!\times\!32$ dataset is more difficult than the ImageNet dataset and all the images are resized to $32\!\times\!32$ for training and test~\cite{denoord2016pixel}.

\begin{table}[!t]
 \centering
 \caption{The means and standard deviations of the accuracies (\%) on the Cifar-$10$ and the Cifar-$100$ datasets, and the $p$-values ($p$) of the Student's $t$-test between the accuracies of the proposed method and the other methods. Note that the best and the second best results of each base model are marked in \textbf{bold} and \underline{\emph{italic}} fonts, respectively. The significance level $\alpha$ is $0.05$. Notation: ``*'': $p<0.05$.}
 \resizebox{\linewidth}{!}{
    \begin{tabular}{lcc|cc}
    \toprule
    Model & Cifar-$10$ & $p$-value & Cifar-$100$ & $p$-value \\
    \midrule
    VGG$16$ (baseline) & $93.90\pm0.10$ & * & $73.64\pm0.14$ & * \\
    VGG$16$+SENet ($2018$) & $94.04\pm0.09$ & * & $73.86\pm0.03$ & * \\
     VGG$16$+NLNet ($2018$) &  $93.95\pm0.34$ &  * &  $73.86\pm0.07$ &  * \\
    VGG$16$+CBAM-S ($2018$) & $\underline{\textit{94.12}}\pm\underline{\textit{0.10}}$ & * & $\underline{\textit{74.31}}\pm\underline{\textit{0.06}}$ & * \\
    VGG$16$+ECA ($2020$) & $94.11\pm0.07$ & * & $73.65\pm0.18$ & * \\
    VGG$16$+GCT ($2020$) & $\underline{\textit{94.12}}\pm\underline{\textit{0.12}}$ & * & $73.67\pm0.23$ & * \\
    VGG$16$+GPCA (ours) & $\boldsymbol{94.51\pm0.03}$ & N/A & $\boldsymbol{74.65\pm0.01}$ & N/A \\
    \midrule
    ResNet$50$ (baseline) & $93.52\pm0.05$ & * & $71.88\pm0.06$ & * \\
    ResNet$50$+SENet ($2018$) & $94.16\pm0.05$ & * & $72.64\pm0.05$ & * \\
     ResNet$50$+NLNet ($2018$) &  $93.68\pm0.15$ &  * &  $71.31\pm1.65$ &  * \\
    ResNet$50$+BAM ($2018$) & $94.14\pm0.04$ & * & $\underline{\textit{73.36}}\pm\underline{\textit{0.09}}$ & * \\
    ResNet$50$+CBAM ($2018$) & $94.06\pm0.03$ & * & $73.01\pm0.06$ & * \\
    ResNet$50$+CBAM-S ($2018$) & $94.00\pm0.05$ & * & $73.29\pm0.04$ & * \\
    ResNet$50$+SRP ($2019$) & $94.11\pm0.01$ & * & $72.25\pm0.14$ & * \\
    ResNet$50$+OA ($2019$) & $94.09\pm0.06$ & * & $72.73\pm0.10$ & * \\
    ResNet$50$+ECA ($2020$) & $93.92\pm0.21$ & * & $72.38\pm0.20$ & * \\
    ResNet$50$+GCT ($2020$) & $\underline{\textit{94.37}}\pm\underline{\textit{0.25}}$ & * & $72.91\pm0.11$ & * \\
    ResNet$50$+GPCA (ours) & $\boldsymbol{94.93\pm0.01}$ & N/A & $\boldsymbol{73.93\pm0.02}$ & N/A \\
    \midrule
    DenseNet$40$ (baseline) & $93.02\pm0.07$ & * & $71.47\pm0.07$ & * \\
    DenseNet$40$+SENet ($2018$) & $93.22\pm0.13$ & * & $\underline{\textit{72.52}}\pm\underline{\textit{0.05}}$ & * \\
     DenseNet$40$+NLNet ($2018$) &  $93.47\pm0.22$ &  * &  $71.38\pm0.06$ &  * \\
    DenseNet$40$+BAM ($2018$) & $\underline{\textit{93.63}}\pm\underline{\textit{0.08}}$ & * & $71.67\pm0.15$ & * \\
    DenseNet$40$+CBAM ($2018$) & $93.19\pm0.14$ & * & $71.93\pm0.07$ & * \\
    DenseNet$40$+CBAM-S ($2018$) & $93.43\pm0.18$ & * & $71.84\pm0.12$ & * \\
    DenseNet$40$+SRP ($2019$) & $93.20\pm0.06$ & * & $71.85\pm0.01$ & * \\
    DenseNet$40$+OA ($2019$) & $93.11\pm0.09$ & * & $71.61\pm0.09$ & * \\
    DenseNet$40$+ECA ($2020$) & $93.37\pm0.08$ & * & $72.13\pm0.06$ & * \\
    DenseNet$40$+GCT ($2020$) & $93.61\pm0.21$ & * & $72.41\pm0.22$ & * \\
    DenseNet$40$+GPCA (ours) & $\boldsymbol{94.16\pm0.03}$ & N/A & $\boldsymbol{73.02\pm0.19}$ & N/A \\
    \midrule
     DenseNet-BC$100$ (baseline) &  $94.38\pm0.16$ &  * &  $76.99\pm0.25$ &  * \\
     DenseNet-BC$100$+SENet ($2018$) &  $93.80\pm0.07$ &  * &  $77.30\pm0.12$ &  * \\
     DenseNet-BC$100$+NLNet ($2018$) &  $94.61\pm0.24$ &  * &  $78.10\pm0.16$ &  * \\
     DenseNet-BC$100$+BAM ($2018$) &  $94.24\pm0.17$ &  * &  $78.15\pm0.31$ &  * \\
     DenseNet-BC$100$+CBAM ($2018$) &  $94.39\pm0.04$ &  * &  $77.54\pm0.34$ &  * \\
      DenseNet$40$+CBAM-S ($2018$) &  $94.49\pm0.10$ &  * &  $77.52\pm0.21$ &  * \\
      DenseNet$40$+SRP ($2019$) &  $94.53\pm0.08$ &  * &  $\underline{\textit{78.31}}\pm\underline{\textit{0.21}}$ &  * \\
      DenseNet$40$+OA ($2019$) &  $94.47\pm0.20$ &  * &  $77.05\pm0.05$ &  *\\
     DenseNet-BC$100$+ECA ($2020$) &  $\underline{\textit{94.66}}\pm\underline{\textit{0.11}}$ &  * &  $78.06\pm0.31$ &  * \\
     DenseNet-BC$100$+GCT ($2020$) &  $94.62\pm0.01$ &  * &  $78.27\pm0.21$ &  * \\
     DenseNet-BC$100$+GPCA (ours) &  $\boldsymbol{95.41\pm0.25}$ &  N/A &  $\boldsymbol{78.81\pm0.18}$ &  N/A \\
    \bottomrule
    \end{tabular}}
 \label{tab:cifar}
 \vspace{-6mm}
\end{table}

\vspace{-4mm}
\subsection{Implementation Details}\label{ssec:implement}

For the Cifar-$10$ and the Cifar-$100$ datasets, VGG$16$~\cite{simonyan2015very}, ResNet$50$~\cite{he2016deep}, and DenseNet$40$~\cite{huang2017densely} models were used as the base models for the proposed GPCA and other referred modules. Adopting the stochastic gradient descent (SGD) optimizer, we trained each model $300$ epochs with batch size of $256$, and the initial learning rates were set as $0.1$ and decayed by a factor of $10$ at the $150^{th}$ and the $225^{th}$ epochs. The momentum and the weight decay values were kept as $0.9$ and $5\times10^{-4}$, respectively. All the methods were conducted three times with random initialization and the means and the standard deviations of the classification accuracies are reported.

For the~\emph{mini}ImageNet and the ImageNet-$32\!\times\!32$ datasets, the VGG$16$, the ResNet$18$, and the ResNet$34$ models were used as the base models for all the methods. The settings of the optimizers and the initial learning rates were the same with the other two datasets. We trained each model $300$ and $100$ epochs with batch sizes of $128$ and $256$ on the two datasets, respectively. Meanwhile, the learning rate on the~\emph{mini}ImageNet dataset was decayed by a factor of $10$ at the $150^{th}$ and the $225^{th}$ epochs, which were the $50^{th}$ and the $75^{th}$ epochs on the other dataset. All the methods were also conducted three times with random initialization and the means and the standard deviations of the classification accuracies are reported as well.

\begin{table}[!t]
 \centering
 \caption{The means and standard deviations of test accuracies (acc., \%) on the~\emph{mini}ImageNet dataset, and the $p$-values ($p$) of the Student's $t$-test between the accuracies of the proposed method and the other methods. Note that the best and the second best results of each base model are marked in \textbf{bold} and \underline{\emph{italic}} fonts, respectively. The significance level $\alpha$ is $0.05$. Notation: ``*'': $p<0.05$.}
 \footnotesize 
    \begin{tabular}{lcc}
    \toprule
    Model & Acc.  & $p$-value \\
    \midrule
    VGG$16$ (baseline) & $81.36\pm0.08$ & * \\
    VGG$16$+SENet ($2018$) & $81.66\pm0.08$ & * \\
    VGG$16$+NLNet ($2018$) &  $80.42\pm0.28$ &  * \\
    VGG$16$+BAM ($2018$) & $81.38\pm0.04$ & * \\
    VGG$16$+CBAM ($2018$) & $81.42\pm0.13$ & * \\
    VGG$16$+CBAM-S ($2018$) & $81.62\pm0.04$ & * \\
    VGG$16$+SRP ($2019$) & $\underline{\textit{81.91}}\pm\underline{\textit{0.13}}$ & * \\
    VGG$16$+ECA ($2020$) & $81.74\pm0.15$ & * \\
    VGG$16$+GCT ($2020$) & $81.68\pm0.19$ & * \\
    VGG$16$+GPCA (ours) & $\boldsymbol{82.78\pm0.01}$ & N/A \\
    \midrule
    ResNet$18$ (baseline) & $77.13\pm0.06$ & * \\
    ResNet$18$+SENet ($2018$) & $77.76\pm0.04$ & * \\
    ResNet$18$+NLNet ($2018$) &  $77.64\pm0.19$ &  * \\
    ResNet$18$+BAM ($2018$) & $77.87\pm0.03$ & * \\
    ResNet$18$+CBAM ($2018$) & $77.62\pm0.07$ & * \\
    ResNet$18$+CBAM-S ($2018$) & $\underline{\textit{78.02}}\pm\underline{\textit{0.04}}$ & * \\
    ResNet$18$+SRP ($2019$) & $77.93\pm0.06$ & * \\
    ResNet$18$+OA ($2019$) & $77.49\pm0.06$ & * \\
    ResNet$18$+ECA ($2020$) & $77.72\pm0.07$ & * \\
    ResNet$18$+GCT ($2020$) & $77.60\pm0.19$ & * \\
    ResNet$18$+GPCA (ours) & $\boldsymbol{78.87\pm0.02}$ & N/A \\
    \midrule
    ResNet$34$ (baseline) & $78.08\pm0.12$ & * \\
    ResNet$34$+SENet ($2018$) & $78.44\pm0.02$ & * \\
    ResNet$34$+NLNet ($2018$) &  $78.48\pm0.2$ &  * \\
    ResNet$34$+BAM ($2018$) & $78.23\pm0.08$ & * \\
    ResNet$34$+CBAM ($2018$) & $78.16\pm0.10$ & * \\
    ResNet$34$+CBAM-S ($2018$) & $78.46\pm0.03$ & * \\
    ResNet$34$+SRP ($2019$) & $\underline{\textit{78.99}}\pm\underline{\textit{0.03}}$ & * \\
    ResNet$34$+OA ($2019$) & $78.14\pm0.04$ & * \\
    ResNet$34$+ECA ($2020$) & $78.68\pm0.14$ & * \\
    ResNet$34$+GCT ($2020$) & $78.17\pm0.12$ & * \\
    ResNet$34$+GPCA (ours) & $\boldsymbol{80.14\pm0.02}$ & N/A \\
    \bottomrule
    \end{tabular}
 \label{tab:miniimagenet}
 \vspace{-6mm}
\end{table}

{For the ImageNet dataset, the ResNet$18$ and the ResNet$50$ models were used as the base models. The settings of the optimizers and the initial learning rates were the same with the other datasets. We trained for $100$ epochs with batch sizes of $256$. Meanwhile, the learning rate on the dataset was decayed by a factor of $10$ at the $50^{th}$ and the $75^{th}$ epochs. The proposed method was also conducted three times with random initialization and the means and the standard deviations of the classification accuracies are reported as well.}

In addition, the two-sample Student's $t$-tests have been conducted between the accuracies of the proposed method and the other referred methods. We set the significance level $\alpha$ as $0.05$,~\emph{i.e.}, statistically significant difference can be found when $p$-value ($p$) of the Student's $t$-test is smaller than $0.05$; otherwise, there is no statistically significant difference.

\begin{table}[!t]
 \centering
 \caption{The means and standard deviations of both top-$1$ and top-$5$ test accuracies (acc., \%) on the ImageNet-$32\times32$ dataset, and the $p$-values ($p$) of the Student's $t$-test between the accuracies of the proposed method and the other methods. Note that the best and the second best results of each base model are marked in \textbf{bold} and \underline{\emph{italic}} fonts, respectively. The significance level $\alpha$ is $0.05$. Notation: ``*'': $p<0.05$.}
 \resizebox{\linewidth}{!}{
    \begin{tabular}{lcc|cc}
    \toprule
    Model & Top-$1$ acc. & $p$-value & Top-$5$ acc. & $p$-value \\
    \midrule
    VGG$16$ (baseline) & $39.70\pm0.02$ & * & $62.66\pm0.03$ & * \\
    VGG$16$+SENet ($2018$) & $39.82\pm0.11$ & * & $62.99\pm0.10$ & * \\
    VGG$16$+NLNet ($2018$) &  $41.39\pm0.22$ &  * &  $64.11\pm0.18$ &  * \\
    VGG$16$+CBAM-S ($2018$) & $39.56\pm0.08$ & * & $62.84\pm0.06$ & * \\
    VGG$16$+ECA ($2020$) & $\underline{\textit{41.89}}\pm \underline{\textit{0.06}}$ & * & $\underline{\textit{64.45}}\pm \underline{\textit{0.18}}$ & * \\
    VGG$16$+GCT ($2020$) & $41.79\pm 0.03$ & * & $\underline{\textit{64.45}}\pm \underline{\textit{0.32}}$ & * \\
    VGG$16$+GPCA (ours) & $\boldsymbol{42.58\pm0.01}$ & N/A      & $\boldsymbol{65.65\pm0.07}$ & N/A \\
    \midrule
    ResNet$18$ (baseline) & $47.75\pm0.07$ & * & $72.70\pm0.13$ & * \\
    ResNet$18$+SENet ($2018$) & $47.89\pm0.09$ & * & $72.98\pm0.02$ & * \\
    ResNet$18$+NLNet ($2018$) &  $47.6\pm0.43$ &  * &  $72.53\pm0.27$ &  * \\
    ResNet$18$+BAM ($2018$) & $47.90\pm0.07$ & * & $73.08\pm0.03$ & * \\
    ResNet$18$+CBAM ($2018$) & $48.48\pm0.08$ & * & $73.25\pm0.03$ & * \\
    ResNet$18$+CBAM-S ($2018$) & $48.31\pm0.13$ & * & $73.37\pm0.09$ & * \\
    ResNet$18$+SRP ($2019$) & $\underline{\textit{48.84}}\pm\underline{\textit{0.04}}$ & * & $\underline{\textit{73.74}}\pm\underline{\textit{0.04}}$ & * \\
    ResNet$18$+OA ($2019$) & $48.62\pm0.08$ & * & $73.21\pm0.05$ & * \\
    ResNet$18$+ECA ($2020$) & $48.44\pm 0.09$ & * & $73.11\pm 0.08$ & * \\
    ResNet$18$+GCT ($2020$) & $48.70\pm 0.12$ & * & $72.95\pm 0.03$ & * \\
    ResNet$18$+GPCA (ours) & $\boldsymbol{49.58\pm0.02}$ & N/A & $\boldsymbol{75.07\pm0.06}$ & N/A \\
    \midrule
    ResNet$34$ (baseline) & $49.82\pm0.10$ & * & $74.34\pm0.02$ & * \\
    ResNet$34$+SENet ($2018$) & $50.30\pm0.04$ & * & $74.89\pm0.21$ & * \\
    ResNet$18$+NLNet ($2018$) &  $49.46\pm0.13$ &  * &  $74.65\pm0.32$ &  * \\
    ResNet$34$+BAM ($2018$) & $\underline{\textit{50.70}}\pm\underline{\textit{0.01}}$ & * & $74.79\pm0.03$ & * \\
    ResNet$34$+CBAM ($2018$) & $50.28\pm0.05$ & * & $\underline{\textit{74.89}}\pm\underline{\textit{0.03}}$ & * \\
    ResNet$34$+CBAM-S ($2018$) & $50.65\pm0.09$ & * & $75.07\pm0.08$ & * \\
    ResNet$34$+SRP ($2019$) & $50.24\pm0.10$ & * & $74.73\pm0.07$ & * \\
    ResNet$34$+OA ($2019$) & $50.04\pm0.06$ & * & $74.88\pm0.03$ & * \\
    ResNet$34$+ECA ($2020$) & $50.33\pm 0.07$ & * & $74.59\pm 0.23$ & * \\
    ResNet$34$+GCT ($2020$) & $50.23\pm 0.06$ & * & $74.47\pm 0.14$ & * \\
    ResNet$34$+GPCA (ours) & $\boldsymbol{51.24\pm0.02}$ & N/A & $\boldsymbol{75.79\pm0.04}$ & N/A \\
    \bottomrule
    \end{tabular}}
 \label{tab:imagenet32x32}
 \vspace{-2mm}
\end{table}

All the attention modules were added in each residual block in the ResNets, but they were only applied after the last convolutional layer of the VGG$16$ model. Note that in a VGG$16$ model, the output size of feature maps of the last convolutional layer was $512\times1\times1$ while the size of input images was $3\times32\times32$ (\emph{i.e.}, the Cifar-$10$, the Cifar-$100$, and the ImageNet-$32\!\times\!32$ datasets), and the BAM, the CBAM, and the SRP cannot work in this case. However, the channel attention of the CBAM can be separately applied in a VGG$16$ model, named as CBAM-S. In addition, the OA needs to be used in multiple blocks and cannot be implemented in the VGG$16$ model on all the datasets either.

\vspace{-4mm}
\subsection{Performance on Cifar-$\boldsymbol{10}$/-$\boldsymbol{100}$ Datasets}\label{ssec:cifar}

From Table~\ref{tab:cifar}, the proposed GPCA module performs the best among all the methods with the VGG$16$, the ResNet$50$, and the DenseNet$40$ models on both Cifar-$10$ and Cifar-$100$ datasets, respectively. Firstly, on the Cifar-$10$ dataset, the proposed GPCA module achieves the classification accuracies at $94.51\%$, $94.93\%$, and $94.16\%$ based on the VGG$16$, the ResNet$50$, the DenseNet$40$ models, respectively. Meanwhile, the proposed GPCA module with different base models exceeds their baseline models, at about $0.6\%$, $1.4\%$, $1.1\%$ improvements, respectively. Moreover, the proposed GPCA module with the VGG$16$ model outperforms the CBAM-S and the GCT module ($94.12\%$ of both), the two best referred methods, at about $0.4\%$, while the proposed GPCA module with the ResNet$50$ model surpasses the GCT module with the ResNet$50$ model ($94.37\%$) more than $0.5\%$. The proposed GPCA module with the DenseNet$40$ model also achieves more than $0.5\%$ improvement than the BAM module with the DenseNet$40$ model.

Similarly, on the Cifar-$100$ dataset, the proposed GPCA module achieves the classification accuracies at $74.65\%$, $73.93\%$, $73.02\%$ on the VGG$16$, the ResNet$50$, and the DenseNet$40$ base models, respectively. It performs the best among all the methods as shown in Table~\ref{tab:cifar}. The GPCA-based VGG$16$, ResNet$50$, and DenseNet$40$ models have significant improvements compared with the corresponding base models at more than $1\%$, $2\%$, and $1.5\%$, respectively. In addition, they also gain about $0.3\%$, $0.6\%$, and $0.5\%$ improvements compared with the best referred models,~\emph{i.e.}, the CBAM-S module with the VGG$16$ ($74.31\%$), the BAM module with the ResNet$50$ ($73.36\%$), and the SENet module with the DenseNet$40$ ($72.52\%$) models, respectively.

According to the $p$-values of Student's $t$-test between the proposed method and the other methods in Table~\ref{tab:cifar}, the results obtained by GPCA are all statistically significant.

\vspace{-4mm}
\subsection{Performance on~\emph{Mini}ImageNet Dataset}\label{ssec:miniimagenet}

In Table~\ref{tab:miniimagenet}, the proposed GPCA module obtained the best results with VGG$16$, ResNet$18$ and ResNet$34$ models among the referred methods on the \emph{mini}ImageNet dataset, which achieves the classification accuracies at $82.78\%$, $78.87\%$, and $80.14\%$, respectively. Applying VGG$16$ as the base model, the classification accuracy of the GPCA with the VGG$16$ model is larger than $82.5\%$. At the meantime, the accuracies obtained by the other referred models are all smaller than $82\%$. It outperforms the base model at about $1.4\%$ and the best referred method, the SRP with the VGG$16$ model, at more than $0.8\%$. Furthermore, the ResNet-based GPCA models also have improvement on classification accuracies on the~\emph{mini}ImageNet dataset. Compared with their base models, the GPCA with the ResNet$18$ and the ResNet$34$ models increase their accuracies about $1.7\%$ and $2.1\%$, respectively, which are considerable performance improvements. Meanwhile, they respectively surpass the CBAM-S with the ResNet$18$ ($78.02\%$) and the SRP with the ResNet$34$ ($78.99$) models more than $0.8\%$ and $1.1\%$, respectively.

Since the $p$-values in Table~\ref{tab:miniimagenet} are all smaller than $0.05$, the GPCA achieves statistically significant improvement on classification performance on the \emph{mini}ImageNet dataset.

\begin{table}[!t]
  \centering
  \caption{The means and standard deviations of both top-$1$ and top-$5$ test accuracies (acc., \%) on the ImageNet dataset, and the $p$-values ($p$) of the Student's $t$-test between the accuracies of the proposed method and the other methods. Note that the best and the second best results of each base model are marked in \textbf{bold} and \underline{\emph{italic}} fonts, respectively. The significance level $\alpha$ is $0.05$. Notation: ``*'': $p<0.05$ and  ``$\dagger$'': the results in the row are from the original papers.}
  \resizebox{\linewidth}{!}{
    \begin{tabular}{lcccc}
    \toprule
     Model &  Top-$1$ acc. &  $p$-value &  Top-$5$ acc. &  $p$-value \\
    \midrule
     ResNet$18$ (baseline)$\dagger$ &  $70.40$ &  *     &  $89.45$ &  * \\
     ResNet$18$+SENet ($2018$)$\dagger$ &  $70.59$ &  *     &  $89.78$ &  * \\
     ResNet$18$+BAM ($2018$)$\dagger$ &  $\underline{\textit{71.12}}$ &  *     &  $\underline{\textit{89.99}}$ &  * \\
     ResNet$18$+CBAM ($2018$)$\dagger$ &  $70.73$ &  *     &  $89.91$ &  * \\
     ResNet$18$+GPCA (ours) &  $\boldsymbol{71.61\pm0.10}$ &  N/A   &  $\boldsymbol{90.90\pm0.13}$ &  N/A \\
    \midrule
     ResNet$50$ (baseline)$\dagger$ &  $75.20$ &  *     &  $92.52$ &  * \\
     ResNet$50$+SENet ($2018$)$\dagger$ &  $76.71$ &  *     &  $93.31$ &  * \\
     ResNet$50$+BAM ($2018$)$\dagger$ &  $75.98$ &  *     &  $92.82$ &  * \\
     ResNet$50$+CBAM ($2018$)$\dagger$ &  $77.34$ &  *     &  $93.66$ &  * \\
     ResNet$50$+SRP ($2019$)$\dagger$ &  $77.42$ &  *     &  $93.79$ &  * \\
     ResNet$50$+ECA ($2020$)$\dagger$ &  $\underline{\textit{77.48}}$ &  *     &  $93.68$ &  * \\
     ResNet$50$+GCT ($2020$)$\dagger$ &  $77.30$ &  *     &  $\underline{\textit{93.70}}$ &  * \\
     ResNet$50$+GPCA (ours) &  $\boldsymbol{77.57\pm0.07}$ &  N/A   &  $\boldsymbol{93.97\pm0.06}$ &  N/A \\
    \bottomrule
    \end{tabular}}
  \label{tab:imagenet}
  \vspace{-6mm}
\end{table}

\vspace{-4mm}
\subsection{Performance on ImageNet-$\boldsymbol{32\!\times\!32}$ Dataset}\label{ssec:imagenet32x32}

In Table~\ref{tab:imagenet32x32}, the proposed GPCA modules with the VGG$16$, the ResNet$18$, and the ResNet$34$ models perform the best on the ImageNet-$32\!\times\!32$ dataset and respectively achieve the top-$1$ accuracies of $42.58\%$, $49.58\%$, and $51.24\%$, and the top-$5$ accuracies of $65.65\%$, $75.07\%$, and $75.79\%$. Compared with their corresponding base models, the models based on the proposed GPCA module improve the accuracies by a large margin. Applying VGG$16$ as the base model, the classification accuracy of the GPCA-based model outperforms the second best one,~\emph{i.e.}, the SENet module with the VGG$16$ model, at $2.8\%$ on top-$1$ and $2.7\%$ on top-$5$, respectively. Meanwhile, with the ResNet$18$ base model, the GPCA-based model has considerable performance improvements compared with the base model and the second best model as well. Compared with the base model, the proposed GPCA with the ResNet$18$ model achieves $1.8\%$ and $2.4\%$ improvements on top-$1$ and top-$5$ accuracies, respectively. Moreover, it improves the accuracies at $0.7\%$ on top-$1$ and $1.3\%$ on top-$5$ than the best referred model,~\emph{i.e.}, the SRP module with the ResNet$18$ model. Meanwhile, with the ResNet$34$ model, the proposed GPCA module achieves the best accuracies on both top-$1$ and top-$5$. 

Given that the $p$-values in Table~\ref{tab:imagenet32x32} are smaller than $0.05$, the GPCA achieves statistically significant performance improvement on the ImageNet-$32\!\times\!32$ dataset.

\begin{table}[!t]
  \centering
  \caption{Ablation studies. The means and standard deviations of top-$1$ test accuracies (\%) on all the datasets. ``GPCA w/o prior'' means the GPCA with the Gaussian process prior removed. ``GPCA-fixed'' means the GPCA with fixed parameter $\Theta$. Note that the best results of each base model are marked in \textbf{bold}, respectively.}
  \footnotesize 
    \begin{tabular}{lcc}
    \toprule
    Model & Cifar-$10$ & Cifar-$100$ \\
    \midrule
    VGG$16$ (baseline) & $93.90\pm0.10$ & $73.64\pm0.14$ \\
    VGG$16$+GPCA w/o prior & $94.21\pm0.03$ & $73.95\pm0.23$ \\
    VGG$16$+GPCA-fixed & $94.50\pm0.02$ & $74.59\pm0.08$ \\
    VGG$16$+GPCA & $\boldsymbol{94.51\pm0.03}$ & $\boldsymbol{74.65\pm0.01}$ \\
    \midrule
    ResNet$50$ (baseline) & $93.52\pm0.05$ & $71.88\pm0.06$ \\
    ResNet$50$+GPCA w/o prior & $94.31\pm0.13$ & $72.36\pm0.20$ \\
    ResNet$50$+GPCA-fixed & $94.87\pm0.11$ & $73.84\pm0.05$ \\
    ResNet$50$+GPCA & $\boldsymbol{94.93\pm0.01}$ & $\boldsymbol{73.93\pm0.02}$ \\
    \midrule
    \midrule
    Model & \emph{Mini}ImageNet & ImageNet-$32\times32$ \\
    \midrule
    VGG$16$ (baseline) & $81.36\pm0.08$ & $39.70\pm0.02$ \\
    VGG$16$+GPCA w/o prior & $81.30\pm0.37$ & $41.36\pm0.16$ \\
    VGG$16$+GPCA-fixed & $82.52\pm0.15$ & $42.15\pm0.11$ \\
    VGG$16$+GPCA & $\boldsymbol{82.78\pm0.01}$ & $\boldsymbol{42.58\pm0.01}$ \\
    \midrule
    ResNet$18$ (baseline) & $77.13\pm0.06$ & $47.75\pm0.07$ \\
    ResNet$18$+GPCA w/o prior & $77.81\pm 0.24$ & $47.37\pm0.13$ \\
    ResNet$18$+GPCA-fixed & $78.85\pm0.09$ & $49.34\pm0.09$ \\
    ResNet$18$+GPCA & $\boldsymbol{78.87\pm0.02}$ & $\boldsymbol{49.58\pm0.02}$ \\
    \midrule
    ResNet$34$ (baseline) & $78.08\pm0.12$ & $49.82\pm0.10$ \\
    ResNet$34$+GPCA w/o prior & $78.34\pm0.36$ & $50.47\pm0.08$ \\
    ResNet$34$+GPCA-fixed & $79.92\pm0.06$ & $50.65\pm0.05$ \\
    ResNet$34$+GPCA & $\boldsymbol{80.14\pm0.02}$ & $\boldsymbol{51.24\pm0.02}$ \\
    \bottomrule
    \end{tabular}\label{tab:ablation}
    \vspace{-4mm}
\end{table}

\begin{figure*}[!t]
 \centering
 \includegraphics[width=0.84\linewidth]{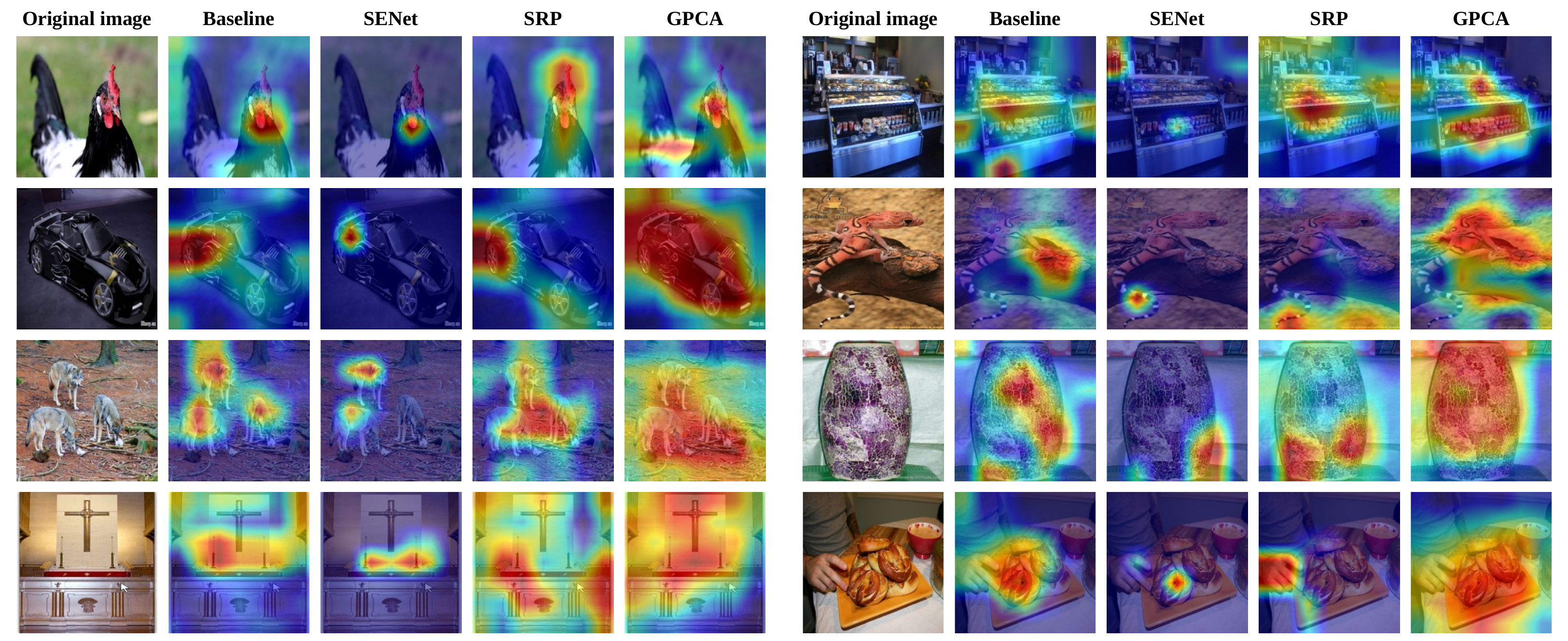}
 \vspace{-3mm}
 \caption{Visualization of feature maps of the last convolutional layer in the VGG$16$ model on the \emph{mini}ImageNet dataset. The proposed GPCA module is compared with the baseline (the VGG$16$ model without any attention modules introduced) and two referred methods (SENet \cite{hu2018squeeze} and SRP \cite{luo2019stochastic}). Red means the highest concentration, while blue is the fully contrary.} \label{fig:visualization}
 \vspace{-6mm}
\end{figure*}

\vspace{-4mm}
\subsection{{Performance on ImageNet Dataset}}\label{ssec:imagenet}

{In Table~\ref{tab:imagenet}, the proposed GPCA modules with the ResNet$18$ and the ResNet$50$ models perform the best on the ImageNet dataset and respectively achieve the top-$1$ accuracies of $71.61\%$ and $77.57\%$, and the top-$5$ accuracies of $90.90\%$ and $93.97\%$. Compared with their corresponding base models, the models based on the proposed GPCA module improve the accuracies by a large margin. Moreover, it improves the accuracies at $0.5\%$ on top-$1$ and $1.0\%$ on top-$5$ than the best referred model,~\emph{i.e.}, the BAM module with the ResNet$18$ model. Meanwhile, with the ResNet$50$ model, the proposed GPCA module achieves the best accuracies on both top-$1$ and top-$5$. The GPCA achieves statistically significant performance improvement on the ImageNet dataset, as the $p$-values in Table~\ref{tab:imagenet} are all smaller than $0.05$.}

\vspace{-4mm}
\subsection{Ablation Studies}\label{ssec:ablation}

Ablation study is conducted and the results are listed in Table~\ref{tab:ablation}. To investigate the effectiveness of the optimization, the classification accuracies of the Gaussian process-removed version (``model\_name+GPCA wo prior'') and the fixed-parameter version (``model\_name+GPCA-fixed'') are reported. For the Gaussian process-removed version, the means and the variances of each element in $\boldsymbol{U}$ are directly optimized under the backpropagation framework. Then, the attention mask vector $\boldsymbol{V}$ will be generated according to Section~\ref{ssec:betaapprox}.

The optimal parameter $\boldsymbol{\Theta}$ of the fixed-parameter version are empirically selected. $\theta_0$, $\theta_2$, and $\theta_3$ are selected from the set $\{0,0.5,1,2\}$ and they cannot be $0$ at the same time. $\theta_1$ is selected from the set $\{2^n\}_{n=-2}^{6}$. For different cases (\emph{i.e.}, different base models and datasets), the optimal choices may not be the same. According to our experience, we set $\theta_1=1$ based on the VGG$16$ model implemented on the Cifar-$10$ and the Cifar-$100$ datasets, $\theta_1=32$ for the ResNet$50$ model on the Cifar-$10$ dataset, $\theta_1=0.5$ for the ResNet$50$ model on the Cifar-$100$ dataset, and $\theta_1=64$ for all the other cases. In all these cases, $\theta_0=1$, $\theta_2=0$, and $\theta_3=1$ are fixed. In summary, $\theta_1$ is sensitive for GP training in different cases, while the other three parameters are less sensitive. Other implementation details are the same as those in Section~\ref{ssec:implement}.

We list means and standard deviations of the classification accuracy, and the $p$-values of the Student's $t$-test in Table~\ref{tab:ablation}. The proposed GPCA achieves statistically significantly better performance than the corresponding base models on all the datasets. Meanwhile, the mean performance of the GPCA is better than that of both ``GPCA w/o prior'' and ``GPCA-fixed'' cases, with smaller standard deviations. This indicates that the Gaussian process prior and the parameter optimization are the essential parts of the GPCA module.

\vspace{-4mm}
\subsection{Visualization}\label{ssec:visualization}

Visualization of the proposed GPCA is shown in Figure~\ref{fig:visualization}. We compare the proposed GPCA method with the baseline (VGG$16$ without any attention modules) and two referred methods on the~\emph{mini}ImageNet dataset. The compared methods include SENet~\cite{hu2018squeeze} (a popular channel attention method) and SRP~\cite{luo2019stochastic} (it achieves the best performance among all the referred methods). We applied the gradient-weighted class activation mapping (Grad-CAM)~\cite{selvaraju2017grad} method to implement the visualization. Eight images and their corresponding visualization are illustrated.

\begin{table*}[!t]
  \centering
  \caption{The means and standard deviations of top-$1$ and top-$5$ localization accuracies (loc. acc., \%), the classification accuracies (cls. acc., \%), and the $p$-values ($p$) of the Student's $t$-test between the accuracies of the proposed method and the other methods on the CUB-$200$-$2011$ dataset. Note that the best results of each base model are marked in \textbf{bold}, respectively. We set the significance level $\alpha$ as $0.05$. Notation: ``-'': no results, ``*'': $p<0.05$. $\dagger$ means the results in the row are from the original papers.}
  \vspace{-2mm}
  \scriptsize
    \begin{tabular}{clcccc|cccc}
    \toprule
    \multirow{2}[1]{*}{Base model} & \multicolumn{1}{c}{\multirow{2}[1]{*}{Method}} & \multicolumn{4}{c|}{Loc. acc.} & \multicolumn{4}{c}{Cls. acc.} \\
          &       & Top-$1$ acc. & $p$-value & Top-$5$ acc. & $p$-value & Top-$1$ acc. & $p$-value & Top-$5$ acc. & $p$-value \\
    \midrule
    \multirow{18}[6]{*}{VGG$16$} & ADL ($2019$)$^{\dagger}$ & $52.36$ & *     & -     & -     & $65.27$ & *     & -     & - \\
          & ACoL ($2018$)$^{\dagger}$ & $45.92$ & *     & $56.51$ & *     & $71.9$ & *     & -     & - \\
          & SPG ($2018$)$^{\dagger}$ & $48.93$ & *     & $57.85$ & *     & $75.5$ &       & $92.1$ &  \\
          & CAM ($2016$) & $42.15\pm0.88$ & *     & $51.04\pm0.92$ & *     & $75.38\pm0.20$ &       & $92.26\pm0.36$ &  \\
          & ADL ($2019$) & $45.30\pm0.26$ & *     & $54.57\pm0.26$ & *     & $73.18\pm0.50$ & *     & $91.97\pm0.37$ & * \\
          & ADL-SE ($2018$) & $52.62\pm0.38$ & *     & $68.27\pm0.29$ & *     & $73.54\pm0.18$ & *     & $91.42\pm0.31$ & * \\
          & ADL-SRP ($2019$) & $52.07\pm0.59$ & *     & $68.11\pm0.58$ & *     & $74.26\pm0.34$ & *     & $90.68\pm0.35$ & * \\
          & ADL-GPCA (ours) & $\boldsymbol{54.97\pm0.35}$ & N/A   & $\boldsymbol{69.68\pm0.24}$ & N/A   & $\boldsymbol{75.62\pm0.17}$ & N/A   & $\boldsymbol{92.77\pm0.32}$ & N/A \\
    \cmidrule{2-10}          &  DANet ($2019$)$^{\dagger}$ &  $52.52$ &       &  $61.96$ &      &  $\boldsymbol{75.40}$ &       &  $92.30$ &  \\
          &  DANet ($2019$) &  $50.96\pm0.52$ &  *     &  $60.85\pm0.50$ &  *     &  $74.81\pm0.42$ &       &  $92.15\pm0.18$ &  \\
          &  DANet-SE ($2018$) &  $51.19\pm0.23$ &  *     &  $61.00\pm0.06$ &  *     &  $74.83\pm0.36$ &       &  $92.22\pm0.30$ &  \\
          &  DANet-SRP ($2019$) &  $50.70\pm0.53$ &  *     &  $60.96\pm0.47$ &  *     &  $74.05\pm0.40$ &  *     &  $92.07\pm0.12$ &  \\
          &  DANet-GPCA (ours) &  $\boldsymbol{52.59\pm0.38}$ &  N/A   &  $\boldsymbol{62.05\pm0.18}$ &  N/A   &  $74.77\pm0.39$ &  N/A   &  $\boldsymbol{92.36\pm0.32}$ &  N/A \\
    \cmidrule{2-10}          &  DGDM ($2020$)$^{\dagger}$ &  $54.34$ &  *     &  -     &  -     &  $\boldsymbol{69.85}$ &       &  -     &  - \\
          &  DGDM ($2020$) &  $52.80\pm0.53$ &  *     &  $67.82\pm0.39$ &  *     &  $69.34\pm0.16$ &  *     &  $90.26\pm0.06$ &  * \\
          &  DGDM-SE ($2018$) &  $53.17\pm0.44$ &  *     &  $67.55\pm0.11$ &  *     &  $69.29\pm0.33$ &  *     &  $90.11\pm0.32$ & * \\
          &  DGDM-SRP ($2019$) &  $53.65\pm0.38$ &  *     &  $68.27\pm0.72$ &  *     &  $69.03\pm0.31$ &  *     &  $90.13\pm0.22$ &  * \\
          &  DGDM-GPCA (ours) &  $\boldsymbol{54.90\pm0.18}$ &  N/A   &  $\boldsymbol{69.80\pm0.09}$ &  N/A   &  $69.67\pm0.26$ &  N/A   &  $\boldsymbol{90.90\pm0.31}$ &  N/A \\
    \midrule
    \multirow{7}[2]{*}{MobileNet} & HaS-$32$ ($2017$)$^{\dagger}$ & $44.67$ & *     & -     & -     & $66.64$ & *     & -     & - \\
          & ADL ($2019$)$^{\dagger}$ & $47.74$ & *     & -     & -     & $70.43$ & *     & -     & - \\
          & CAM ($2016$) & $42.81\pm0.07$ & *     & $53.24\pm0.76$ & *     & $69.17\pm0.86$ & *     & $90.11\pm0.25$ & * \\
          & ADL ($2019$) & $45.09\pm0.23$ & *     & $54.32\pm0.40$ & *     & $72.93\pm0.24$ & *     & $90.96\pm0.33$ & * \\
          & ADL-SE ($2018$) & $48.67\pm0.37$ & *     & $59.07\pm0.41$ & *     & $73.05\pm0.36$ & *     & $91.65\pm0.36$ &  \\
          & ADL-SRP ($2019$) & $48.15\pm0.78$ & *     & $58.61\pm0.58$ & *     & $71.98\pm0.20$ & *     & $90.08\pm0.38$ & * \\
          & ADL-GPCA (ours) & $\boldsymbol{50.51\pm0.40}$ & N/A   & $\boldsymbol{61.28\pm0.23}$ & N/A   & $\boldsymbol{73.92\pm0.25}$ & N/A   & $\boldsymbol{91.97\pm0.12}$ & N/A \\
    \midrule
    \multirow{10}[4]{*}{ResNet$50$} & CAM ($2016$) & $42.82\pm0.74$ & *     & $52.13\pm0.82$ & *     & $74.68\pm0.77$ & *     & $92.89\pm0.18$ & * \\
          & ADL ($2019$) & $45.74\pm0.10$ & *     & $54.45\pm0.14$ & *     & $76.86\pm0.20$ & *     & $91.26\pm0.06$ & * \\
          & ADL-SE ($2018$) & $49.03\pm0.17$ & *     & $58.72\pm0.23$ &       & $77.86\pm0.26$ & *     & $93.95\pm0.07$ &  \\
          & ADL-SRP ($2019$) & $45.55\pm0.25$ & *     & $54.62\pm0.08$ & *     & $76.06\pm0.05$ & *     & $91.83\pm0.07$ & * \\
          & ADL-GPCA (ours) & $\boldsymbol{50.73\pm0.47}$ & N/A   & $\boldsymbol{59.66\pm0.63}$ & N/A   & $\boldsymbol{78.75\pm0.08}$ & N/A   & $\boldsymbol{94.17\pm0.28}$ & N/A \\
    \cmidrule{2-10}          &  DGDM ($2020$)$^{\dagger}$ &  $59.40$ &  *     &  -     &  -     &  $76.20$ &       &  -     &  - \\
          &  DGDM ($2020$) &  $57.87\pm0.34$ &  *     &  $69.71\pm0.33$ &  *     &  $76.23\pm0.23$ &       &  $92.64\pm0.19$ &  \\
          &  DGDM-SE ($2018$) &  $62.25\pm0.43$ &  *     &  $76.29\pm0.20$ &  *     &  $\boldsymbol{76.37\pm0.19}$ &  *     &  $\boldsymbol{92.77\pm0.07}$ &  \\
          &  DGDM-SRP ($2019$) &  $62.33\pm0.24$ &  *     &  $76.38\pm0.36$ &  *     &  $75.84\pm0.34$ &       &  $92.63\pm0.22$ &  \\
          &  DGDM-GPCA (ours) &  $\boldsymbol{63.22\pm0.27}$ &  N/A   &  $\boldsymbol{77.20\pm0.18}$ &  N/A   &  $75.80\pm0.23$ &  N/A   &  $92.73\pm0.20$ &  N/A \\
    \bottomrule
    \end{tabular}
  \label{tab:loccub}
  \vspace{-2mm}
\end{table*}

As shown in Figure~\ref{fig:visualization}, the GPCA method focuses on larger discriminative parts of the objects in the images. The key regions (in red) usually enclose the whole object to benefit classification. At the meantime, the other methods only concentrate on relatively small region in the images.

For example, in the second sample of the left column, a car is in the image. The baseline, the SENet, and the SRP can find parts of the car, while the GPCA can detect the car as a whole. Another example, the fourth sample of the right column, shows multiple pretzels (targets) on the table and in a hand. The baseline and the SENet can find parts of the pretzels, but the SRP merely focuses on the hand. At the meantime, the GPCA concentrates on all the pretzels. These indicate that the GPCA has ability to find more informative channels than the others such that we can integrate them for better classification performance.

\vspace{-4mm}
\section{Experiments on Weakly Supervised Object Localization}\label{sec:wsol}

In this section, we evaluate the performance of the proposed GPCA module on one benchmark dataset for the weakly supervised object localization (WSOL) task. We compare the proposed GPCA module with two attention modules,~\emph{i.e.}, SENet~\cite{hu2018squeeze}, and SRP~\cite{luo2019stochastic}, and state-of-the-art WSOL methods including class activation mapping (CAM)~\cite{zhou2016learning}, attention-based dropout layer (ADL)~\cite{cheo2019attention,choe2020attention}, adversarial complementary learning (ACoL)~\cite{zhang2018adversarial}, hide-and-seek (HaS)~\cite{singh2017hide}, self-produced guidance (SPG)~\cite{zhang2018self}, divergent activation network (DANet)~\cite{xue2019danet}, {and dual-attention guided dropblock module (DGDM)~\cite{yin2020dual}}.

\vspace{-4mm}
\subsection{Dataset}\label{ssec:locdatasets}

CUB-$200$-$2011$~\cite{wah2011cub} dataset is used for the WSOL task. The CUB-$200$-$2011$ dataset contains $200$ categories of birds with $5,994$ images for training and $5,794$ images for test. It is mainly motivated for fine-grained visual classification (FGVC) tasks, while previous researches~\cite{zhou2016learning,cheo2019attention,choe2020attention,zhang2018adversarial,singh2017hide,zhang2018self,xue2019danet} commonly worked on it for WSOL.

\vspace{-4mm}
\subsection{Implementation Details}\label{ssec:locimplement}

VGG$16$, MobileNet~\cite{howard2017mobilenets}, and ResNet$50$ models were used as the base models. Adopting the SGD optimizer, we trained each model $100$ epochs with batch size of $32$, and the initial learning rates were set as $0.01$ and decayed by a factor of $10$ at the $80^{th}$ epoch. The momentum and the weight decay values were kept as $0.9$ and $5\times10^{-4}$, respectively. All the models with the attention modules were fine-tuned on ImageNet-pretrained models and conducted three times with different random initializations. The means and the standard deviations of the localization and the classification accuracies are reported, respectively. A correct localization means that the network can predict a correct class and the intersection over union (IoU) between the estimated box and the ground truth is no smaller than $50\%$. The localization accuracy can present the performance of the network more accurate than the classification accuracy.

In addition, the two-sample Student's $t$-tests have been conducted between the accuracies of the proposed method and the other referred methods that we reimplemented. As the work in WSOL always reports results for one run, we conducted the one-sample Student's $t$-tests between the accuracies of the proposed method and the WSOL methods in these work. We set the significance level $\alpha$ as $0.05$.

{For the ADL- and DGDM-based models, we applied ADL or DGDM after each pooling and strided convolutional layer.} In addition, {for all the models,} we replaced the last pooling layer and the FC layer(s) with two $3\times3$ convolutional layers, one $1\times1$ convolutional layer for mapping output the channel index to the class index, and one GAP for shrinking channels to scalars, following~\cite{xue2019danet}. All the attention modules were only applied before the GAP(s).

\vspace{-3mm}
\subsection{Performance on CUB-$\boldsymbol{200}$-$\boldsymbol{2011}$ Dataset}\label{ssec:cub}

Experimental results in the WSOL task are reported in Table~\ref{tab:loccub}. It can be observed that the proposed GPCA module performs better {on both top-$1$ and top-$5$ localization accuracies} than the SENet and the SRP in all the three base models {with three different WSOL methods as baselines}, respectively, and achieves statistically significant improvement on localization accuracies, instead of the top-$5$ accuracies of the SENet with the ADL in the ResNet$50$ model. Meanwhile, compared with the baseline models, the proposed GPCA module can obtain statistically significant improvement on localization as well. The GPCA module performs better than the referred methods, which indicates that the GPCA module can effectively improve feature representation ability.

\begin{figure*}[!t]
 \centering
 \includegraphics[width=0.84\linewidth]{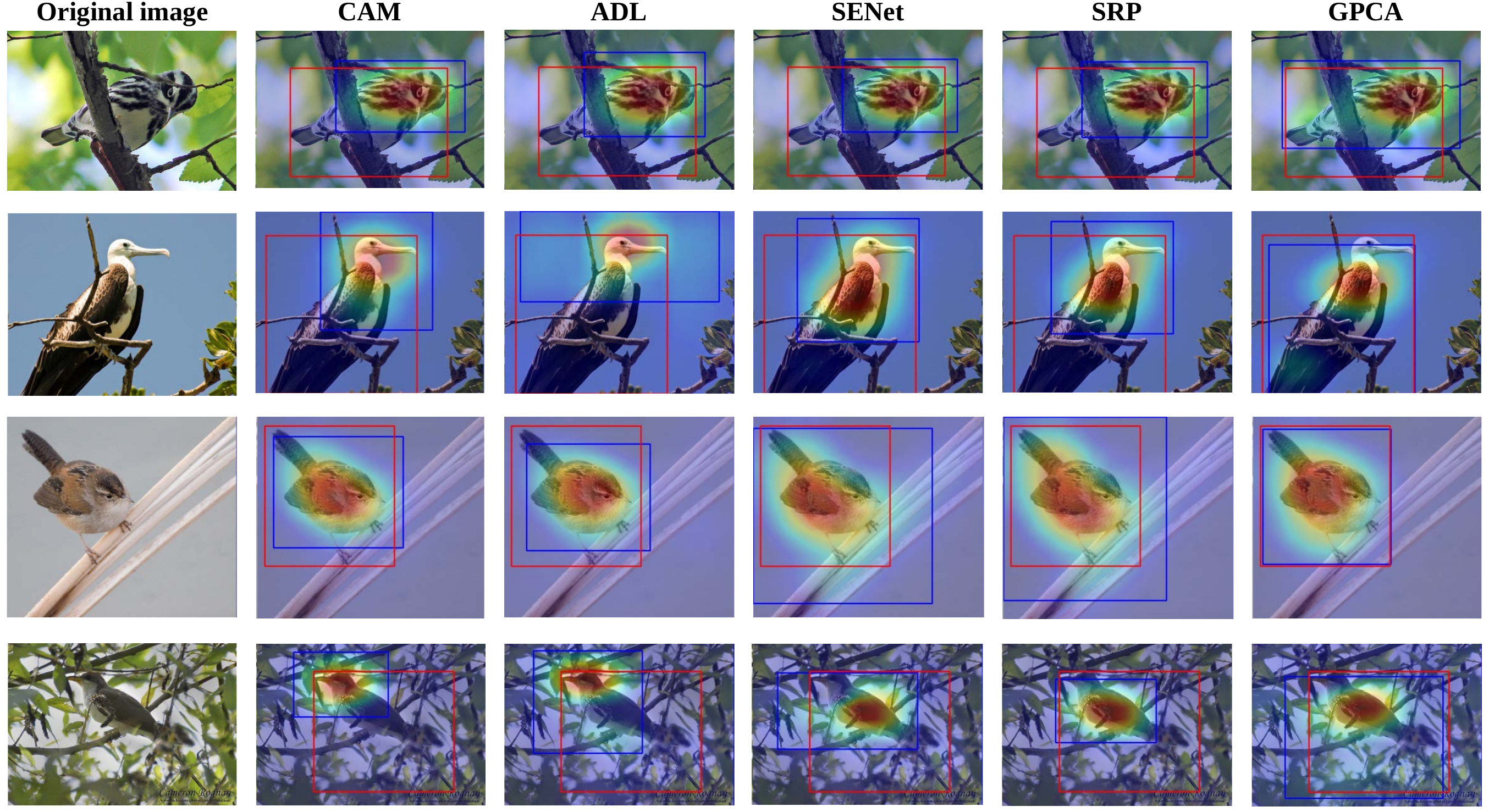}
 \vspace{-3mm}
 \caption{Examples of WSOL results on the CUB-$200$-$2011$ dataset. The proposed GPCA module is compared with two attention modules (SENet and SRP) and two WSOL methods (CAM and ADL). The correctly predicted WSOL results are shown. The blue bounding boxes are the predicted results and the red ones are the ground truths.} \label{fig:locexamples}
 \vspace{-4mm}
\end{figure*}

In addition, we visualize a group of localization results in images from the test set of the CUB-$200$-$2011$ dataset (Figure~\ref{fig:locexamples}). From the Figure~\ref{fig:locexamples}, the proposed GPCA module-based model with the ADL can locate more precise object regions such that the localization performance is improved.

\vspace{-4mm}
\section{Experiments on Object Detection}\label{sec:detection}

In this section, we evaluate the performance of the GPCA on two benchmark datasets for object detection task. We compare the proposed GPCA module with four attention modules,~\emph{i.e.}, SENet~\cite{hu2018squeeze}, SRP~\cite{luo2019stochastic}, ECA~\cite{wang2020ecanet}, and GCT~\cite{yang2020gated}, as all the attention modules can be applied in the detectors.

\vspace{-4mm}
\subsection{Dataset}\label{ssec:detdatasets}

We utilize the PASCAL visual object classes challenge $2007$ (VOC$2007$)~\cite{voc2007} {and common objects in context (COCO)~\cite{lin2014microsoft}} datasets, which are well-known and widely used benchmark datasets, for evaluations of the object detection task. The VOC$2007$ dataset contains $20$ classes with a training set ($5,011$ images) and a test set ($4,952$ images). {The COCO dataset contains $80$ classes with a training set ($118,287$ images) and a test set ($40,670$ images).}

\vspace{-4mm}
\subsection{Implementation Details}\label{ssec:vocim}

On the VOC$2007$ dataset, single shot detector (SSD)~\cite{liu2016ssd} model, which applies the VGG$16$ model as its backbone, was used as the base model for the attention modules. Meanwhile, all the attention modules were only applied after the Conv$5$\_$3$ of the VGG$16$ model for enhancing feature representation. {Two settings, ($1$) training on the VOC$2007$ and testing on it with an ImageNet-pretrained backbone and $300\times300$ input image sizes, ($2$) training on the VOC$2007$ and VOC$2012$ and testing on the VOC$2007$ with a COCO-pretrained backbone and $512\times512$ input image sizes, were undertaken.} For the training process, we followed the original strategy in~\cite{liu2016ssd}. All the models with the attention modules were conducted three times with random initialization. The means of average precision (AP, \%) of each class, and the means and the standard deviations of the mean average precision (mAP, \%) of all the classes are reported, respectively. The intersection over union (IoU) for calculating the AP is set as $50\%$.

{On the COCO dataset, fully convolutional one-stage object detector (FCOS)~\cite{tian2019fcos} and Adaptive Training Sample Selection (ATSS)~\cite{zhang2020bridging} were applied as the base models for the attention modules, respectively. ResNet$50$ was used as the backbone of the two detectors. All the attention modules were cascaded after each outputs of the feature pyramid network (FPN)~\cite{lin2017feature}. Adopting the SGD optimizer, we trained each model twelve epochs with batch size of eight, and the initial learning rates were set as $0.01$ and decayed twice by a factor of $10$ at the eighth and eleventh epochs, respectively. The momentum and the weight decay values were kept as $0.9$ and $5\times10^{-4}$, respectively. Six evaluation metrics were considered including averaged APs for small, medium, and large scale objects and overall at $[50\%,95\%]$ IoU interval with step as $5\%$, and APs at IoU as $50\%$ and $75\%$. We conducted the experiments with MMDetection~\cite{mmdetection}.}

In addition, the two-sample Student's $t$-tests have been conducted between the mAP of the proposed method and the other referred methods that we reimplemented.

\begin{table*}[!t]
  \caption{The means of AP (\%) of each class, the means and standard deviations of mAP (\%) on the VOC$2007$ dataset, and the $p$-values ($p$) of the Student's $t$-test between the mAPs of the proposed method and the other methods. Note that the best and second best results are marked in \textbf{bold} and \underline{\textit{italic}}, respectively. We set the significance level $\alpha$ as $0.05$. Notation: ``*'': $p<0.05$. The class ``Mbike'' means motorbike.}
  \vspace{-2mm}
  \centering
  \resizebox{\textwidth}{!}{
    \begin{tabular}{@{}l@{}|@{}c@{}c@{}c@{}c@{}c@{}c@{}c@{}c@{}c@{}c@{}c@{}c@{}c@{}c@{}c@{}c@{}c@{}c@{}c@{}c@{}|@{}c@{}c@{}}
    \toprule
    \multicolumn{1}{c|}{Method} & Aero & Bike & Bird & Boat & Bottle & Bus & Car & Cat & Chair & Cow & Table & Dog & Horse & Mbike & Person & Plant & Sheep & Sofa & Train & TV & mAP & $p$-value \\
    \midrule
    \multicolumn{23}{c}{Training on VOC$2007$, testing on VOC$2007$, input image size as $300\times300$.} \\
    \midrule
    SSD (baseline) & $74.67$ & $77.94$ & $63.09$ & $61.41$ & $40.19$ & $76.53$ & $80.04$ & $79.49$ & $48.67$ & $70.78$ & $67.42$ & $75.72$ & $81.24$ & $77.22$ & $73.00$ & $40.09$ & $66.70$ & $70.37$ & $78.82$ & $68.78$ & $68.61\pm0.09$ & * \\
    SSD-SENet ($2018$) &\ \ $\boldsymbol{76.76}$\ \  & $79.92$ & $63.83$ & $\underline{\textit{61.57}}$ & $\underline{\textit{41.36}}$ & $78.01$ & $80.66$ & $79.70$ & $48.86$ & $72.33$ & $\boldsymbol{69.14}$\ \ & $\underline{\textit{76.32}}$ & $81.84$ & $77.92$ & $74.01$ & $\underline{\textit{42.28}}$ & $66.41$ & $70.16$ & $79.88$ & $69.13$ & $69.50\pm0.27$ & * \\
    SSD-SRP ($2019$) & $76.24$ & $\boldsymbol{81.04}$\ \  & $62.96$ & $58.94$ & $39.53$ & $76.62$ & $\boldsymbol{84.42}$\ \  & $\boldsymbol{80.19}$\ \  & $50.09$ & $70.77$ & $63.41$ & $73.89$ & $81.71$ & $\underline{\textit{78.22}}$ &$\boldsymbol{75.88}$\ \  & $41.45$ & $\underline{\textit{68.05}}$ & $67.09$ & $\boldsymbol{81.98}$\ \  &$\boldsymbol{71.80}$\ \  & $69.21\pm0.19$ & * \\
    SSD-ECA ($2020$)& $74.99$ & $78.39$ & $62.28$ & $59.55$ & $38.84$ & $78.14$ & $80.80$ & $78.08$ & $48.21$ & $69.47$ & $66.70$ & $74.03$ & $\underline{\textit{82.06}}$ & $76.62$ & $73.76$ & $41.74$ & $\boldsymbol{68.13}$ & $67.42$ & $80.91$ & $68.85$ & $68.45\pm0.92$ & * \\
    SSD-GCT ($2020$) & $75.66$ & $79.37$ & $\underline{\textit{65.51}}$ & $61.14$ & $40.88$ & $\underline{\textit{80.42}}$ & $\underline{\textit{81.74}}$ & $79.74$ & $\boldsymbol{51.24}$\ \  & $\underline{\textit{72.38}}$ & $68.13$ & $75.93$ & $80.39$ & $77.36$ & $74.05$ & $\boldsymbol{43.07}$\ \  & $67.25$ & $\underline{\textit{70.60}}$ & $79.66$ & $70.42$ & $\underline{\textit{69.75}}\pm\underline{\textit{0.22}}$ & * \\
    SSD-GPCA (ours)\ \  & $\underline{\textit{76.51}}$ & $\underline{\textit{80.67}}$ & $\boldsymbol{66.19}$\ \ & $\boldsymbol{62.08}$\ \ & $\boldsymbol{41.97}$\ \ & $\boldsymbol{80.55}$\ \ & $81.06$ & $\underline{\textit{79.89}}$ & $\underline{\textit{50.25}}$ & $\boldsymbol{74.50}$\ \  & $\underline{\textit{68.45}}$ & $\boldsymbol{78.42}$\ \  & $\boldsymbol{82.85}$\ \  & $\boldsymbol{78.76}$\ \  & $\underline{\textit{74.36}}$ & $42.09$ & $67.78$ & $\boldsymbol{70.99}$\ \  & $\underline{\textit{81.13}}$ & $\underline{\textit{71.24}}$ & \ \ $\boldsymbol{70.49\pm0.38}$ & N/A \\
    \midrule
    \multicolumn{23}{c}{Training on VOC$2007$+VOC$2012$+COCO, testing on VOC$2007$, input image size as $512\times512$.} \\
    \midrule
     SSD (baseline) &  $86.57$ &  $86.27$ &  $81.00$ &  $\underline{\textit{77.63}}$ &  $62.97$ &  $86.30$ &  $87.33$ &  $87.50$ &  $67.83$ &  $87.17$ &  $76.77$ &  $85.70$ &  $87.63$ &  $85.07$ &  $82.67$ &  $\boldsymbol{58.30}$ &  $83.30$ &  $78.67$ &  $86.27$ &  $79.50$ &  $80.72\pm0.15$ &  * \\
     SSD-SENet ($2018$) &  $87.83$ &  $\underline{\textit{87.67}}$ &  $\underline{\textit{82.87}}$ &  $76.40$ &  $\boldsymbol{64.23}$ &  $87.27$ &  $88.60$ &  $89.13$ &  $\underline{\textit{68.00}}$ &  $87.30$ &  $76.57$ &  $\underline{\textit{86.47}}$ &  $88.23$ &  $85.97$ &  $83.37$ &  $57.77$ &  $84.00$ &  $78.97$ &  $\underline{\textit{87.47}}$ &  $81.00$ &  $\underline{\textit{81.46}}\pm\underline{\textit{0.06}}$ &  * \\
     SSD-SRP ($2019$) &  $87.63$ &  $87.53$ &  $81.23$ &  $77.03$ &  $\underline{\textit{63.27}}$ &  $\underline{\textit{87.43}}$ &  $\boldsymbol{88.80}$ &  $88.93$ &  $67.27$ &  $\underline{\textit{87.37}}$ &  $77.53$ &  $86.07$ &  $88.23$ &  $\boldsymbol{86.33}$ &  $\boldsymbol{83.57}$ &  $58.17$ &  $\boldsymbol{84.57}$ &  $78.80$ &  $86.47$ &  $80.80$ &  $81.35\pm0.07$ &  * \\
     SSD-ECA ($2020$) &  $87.57$ &  $87.43$ &  $81.83$ &  $77.17$ &  $62.13$ &  $86.73$ &  $\underline{\textit{88.77}}$ &  $89.13$ &  $67.87$ &  $87.03$ &  $76.93$ &  $\underline{\textit{86.47}}$ &  $\boldsymbol{88.60}$ &  $86.23$ &  $83.47$ &  $57.30$ &  $83.40$ &  $78.93$ &  $87.17$ &  $\underline{\textit{81.17}}$ &  $81.27\pm0.19$ &  * \\
     SSD-GCT ($2020$) &  $\underline{\textit{88.00}}$ &  $86.70$ &  $82.20$ &  $76.87$ &  $62.87$ &  $87.23$ &  $\underline{\textit{88.77}}$ &  $\underline{\textit{89.30}}$ &  $\boldsymbol{68.07}$ &  $86.97$ &  $\boldsymbol{78.00}$ &  $86.33$ &  $88.20$ &  $\underline{\textit{86.27}}$ &  $\underline{\textit{83.50}}$ &  $56.73$ &  $84.07$ &  $\underline{\textit{79.10}}$ &  $86.60$ &  $80.63$ &  $81.32\pm0.12$ &  * \\
     SSD-GPCA (ours) &  $\boldsymbol{88.77}$ &  $\boldsymbol{88.13}$ &  $\boldsymbol{82.88}$ &  $\boldsymbol{79.22}$ &  $62.51$ &  $\boldsymbol{88.17}$ &  $88.74$ &  $\boldsymbol{89.69}$ &  $67.92$ &  $\boldsymbol{87.92}$ &  $\underline{\textit{77.66}}$ &  $\boldsymbol{86.71}$ &  $\underline{\textit{88.35}}$ &  $86.12$ &  $83.44$ &  $\underline{\textit{58.23}}$ &  $\underline{\textit{84.43}}$ &  $\boldsymbol{79.31}$ &  $\boldsymbol{88.18}$ &  $\boldsymbol{82.10}$ &  $\boldsymbol{81.92\pm0.11}$ &  N/A \\
    \bottomrule
    \end{tabular}}
  \label{tab:voc}
  \vspace{-4mm}
\end{table*}

\begin{figure*}[!t]
 \centering
 \includegraphics[width=0.84\linewidth]{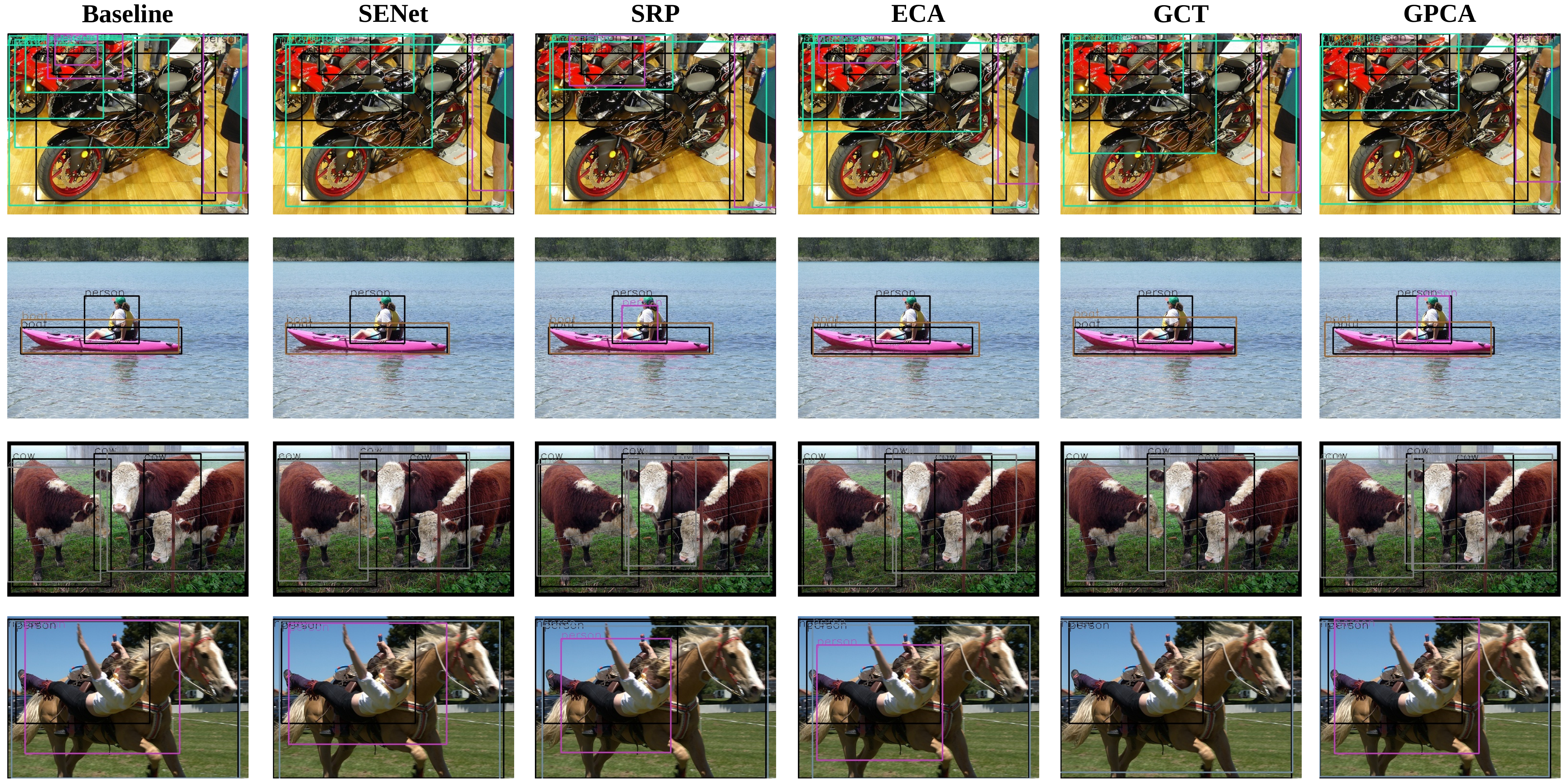}
 \vspace{-3mm}
 \caption{Examples of detection results on the VOC$2007$ dataset. The proposed GPCA module is compared with the baseline (the SSD model without any attention modules introduced) and four referred methods (SENet, SRP, ECA, and GCT). The correct predicted detection results with IoU higher than $0.5$ are shown. Each color corresponds to an object category (green: motor bike; pink: person; brown: boat; grey: cow; light blue: horse) and the black bounding boxes are the ground truths.} \label{fig:detexamples}
 \vspace{-4mm}
\end{figure*}

\vspace{-4mm}
\subsection{Performance on VOC$\boldsymbol{2007}$ Dataset}\label{ssec:voc}

The experimental results of the VOC$2007$ dataset are shown in Table~\ref{tab:voc}. {When training on VOC$2007$ only}, the proposed GPCA module achieves statistically significant improvement on mAP at $70.49\%$ on average and outperforms the second best one,~\emph{i.e.}, the GCT module ($69.75\%$), by $0.75\%$. For all the $20$ classes, the GPCA module performs best on nine classes and the second best on the other eight classes. For the other three classes, the GPCA module obtains small gaps compared with corresponding best attention modules, and is ranked third, respectively. {When training on both VOC$2007$ and VOC$2012$ with larger input image sizes, the proposed GPCA module can further obtain mAP at $81.92\%$ on average, which also performs statistically significant improvement among the referred methods. Meanwhile, the GPCA module performs best on eleven classes and the second best on the other four classes.} These indicate the significant performance of the GPCA module.

We also illustrate four detection examples which are randomly selected from the test set of the VOC$2007$ dataset in Figure~\ref{fig:detexamples}. For each attention module-based model and the baseline model, the predicted bounding boxes of each class are drawn with different colors. From the Figure~\ref{fig:detexamples}, the proposed GPCA module-based model finds all the objects in the images with high IoU between the predictions and the corresponding ground truth. Although the compared methods can perform well sometimes, they merely detect a part of the objects and predict with low IoU in some cases.

\vspace{-4mm}
\subsection{{Performance on COCO Dataset}}\label{ssec:coco}

{The experimental results of the COCO dataset are shown in Table~\ref{tab:coco}. According to Table~\ref{tab:coco}, the proposed GPCA module can outperform the referred attention modules with the two base models, respectively, on the six evaluation metrics. On the prime challenge metric,~\emph{i.e.}, the AP metric, the GPCA module can achieve $38.50\%$ with the FCOS and $39.20\%$ with the ATSS. Moreover, on most of the evaluation metrics, the proposed GPCA module performs statistically significant improvement compared with the referred methods, except the AP$_{\text{50}}$ with the FCOS and the AP$_{\text{L}}$ with the ATSS. These can also demonstrate the significant effectiveness of the GPCA module.}

\vspace{-4mm}
\section{Experiments on Semantic Segmentation}\label{sec:seg}

{In this section, we evaluate the performance of GPCA for the task of semantic segmentation. We compare the proposed GPCA module with SENet~\cite{hu2018squeeze} and ECA~\cite{wang2020ecanet}.}

\begin{table*}[!t]
  \centering
  \caption{Means and standard deviations of AP (\%) in different cases on the COCO dataset, and the $p$-values ($p$) of the Student's $t$-test between the APs of the proposed method and the other methods. Note that the best and second best results are marked in \textbf{bold} and \underline{\textit{italic}}, respectively. We set the significance level $\alpha$ as $0.05$. Notation: ``*'': $p<0.05$, ``AP'', ``AP$_{\text{S}}$'', ``AP$_{\text{M}}$'', and ``AP$_{\text{L}}$'': averaged AP for overall, small, medium, and large scale objects, respectively, at $[50\%,95\%]$ IoU interval with step as $5\%$, ``AP$_{50}$'' and ``AP$_{75}$'': AP at IoU as $50\%$ and $75\%$, respectively.}
  \vspace{-2mm}
  \resizebox{\textwidth}{!}{
    \begin{tabular}{@{}l@{}c@{}c@{}c@{}c@{}c@{}c@{}c@{}c@{}c@{}c@{}c@{}c@{}}
    \toprule
    \multicolumn{1}{c}{Method} &  AP    &  $p$-value &  AP$_{50}$ &  $p$-value &  AP$_{75}$ &  $p$-value &  AP$_{\text{S}}$ &  $p$-value &  AP$_{\text{M}}$ &  $p$-value &  AP$_{\text{L}}$ &  $p$-value \\
    \midrule
     FCOS (baseline) &  $37.50\pm0.10$ &  *     &  $56.43\pm0.21$ &        &  $40.27\pm0.15$ &  *     &  $21.50\pm0.30$ &  *     &  $41.03\pm0.25$ &  *     &  $48.60\pm0.26$ &  * \\
     FCOS-SE ($2018$) &  $37.50\pm0.22$ &  *     &  $55.90\pm0.16$ &  *     &  $40.78\pm0.26$ &  *     &  $21.40\pm0.33$ &  *     &  $41.53\pm0.13$ &  *     &  $48.48\pm0.38$ &  * \\
     FCOS-ECA ($2020$) &  $\underline{\textit{37.90}}\pm\underline{\textit{0.10}}$ &  *     &  $\underline{\textit{56.77}}\pm\underline{\textit{0.67}}$ &        &  $\underline{\textit{41.07}}\pm\underline{\textit{0.12}}$ &        &  $\underline{\textit{21.63}}\pm\underline{\textit{0.40}}$ &  *     &  $\underline{\textit{41.67}}\pm\underline{\textit{0.12}}$ &  *     &  $\underline{\textit{48.67}}\pm\underline{\textit{0.15}}$ &  * \\
     FCOS-GPCA (ours)\ \ \ &  $\boldsymbol{38.50\pm0.20}$ &  N/A   &  $\boldsymbol{56.80\pm0.35}$ &  N/A   &  $\boldsymbol{41.50\pm0.35}$ &  N/A   &  $\boldsymbol{22.17\pm0.29}$ &  N/A   &  $\boldsymbol{42.23\pm0.15}$ &  N/A   &  $\boldsymbol{49.73\pm0.55}$ &  N/A \\
    \midrule
     ATSS (baseline) &  $38.75\pm0.07$ &  *     &  $56.35\pm0.07$ &  *     &  $41.80\pm0.14$ &  *     &  $22.55\pm0.07$ &  *     &  $42.20\pm0.14$ &  *     &  $48.95\pm0.07$ &  * \\
     ATSS-SE ($2018$) &  $38.95\pm0.07$ &  *     &  $\underline{\textit{57.15}}\pm\underline{\textit{0.07}}$ &  *     &  $\underline{\textit{42.00}}\pm\underline{\textit{0.03}}$ &  *     &  $\underline{\textit{23.00}}\pm\underline{\textit{0.14}}$ &  *     &  $\underline{\textit{42.25}}\pm\underline{\textit{0.07}}$ &  *     &  $\underline{\textit{50.15}}\pm\underline{\textit{0.07}}$ &   \\
     ATSS-ECA ($2020$) &  $\underline{\textit{39.00}}\pm\underline{\textit{0.01}}$ &  *     &  $56.95\pm0.07$ &  *     &  $41.95\pm0.07$ &  *     &  $22.75\pm0.07$ &  *     &  $41.90\pm0.14$ &  *     &  $50.05\pm0.07$ &   \\
     ATSS-GPCA (ours)\ \ &  $\boldsymbol{39.20\pm0.01}$ &  N/A   &  $\boldsymbol{57.55\pm0.07}$ &  N/A   &  $\boldsymbol{42.45\pm0.07}$ &  N/A   &  $\boldsymbol{23.55\pm0.07}$ &  N/A   &  $\boldsymbol{42.70\pm0.02}$ &  N/A   &  $\boldsymbol{50.65\pm0.35}$ &  N/A \\
    \bottomrule
    \end{tabular}}
  \label{tab:coco}
  \vspace{-4mm}
\end{table*}

\begin{table}[!t]
  \centering
  \caption{The means and standard deviations of mIoU (\%) and mACC (\%) for the task of semantic segmentation in different backbones on the CityScapes dataset, the $p$-values ($p$) of the Student's $t$-test between mIoUs or mACCs of the proposed method and the other methods. Note that the best and second best results are marked in \textbf{bold} and \underline{\textit{italic}}, respectively. We set the significance level $\alpha$ as $0.05$. HR$18$ and HR$48$ means HRNetV2-W$18$ and HRNetV2-W$48$, respectively. ``*'' means statistically significant difference. {``$\dagger$'' means the results in the row are from~\cite{mmseg2020} with $80,000$ iterations and batch size of $8$.}}
  \vspace{-2mm}
    \begin{tabular}{@{}l@{}c@{}c@{}c@{}c@{}c@{}}
    \toprule
    \multicolumn{1}{c}{Method} &  Backbone &  mIoU &  $p$-value &  mAcc &  $p$-value \\
    \midrule
    OCR (baseline)$^{\dagger}$ & \multirow{5}[0]{*}{HR$18$} &  $78.56$ &  * &  $86.04$ & * \\
     OCR (baseline) &  &  $78.92\pm0.14$ &  * &  $86.45\pm0.13$ &  * \\
     OCR-SE ($2018$) &       &  $\underline{\textit{79.51}}\pm\underline{\textit{0.25}}$ &  * &  $\underline{\textit{86.72}}\pm\underline{\textit{0.14}}$ &  * \\
     OCR-ECA ($2020$) &       &  $79.47\pm0.29$ &  * &  $86.46\pm0.24$ &  * \\
     OCR-GPCA (ours) &       &  $\boldsymbol{80.09\pm0.22}$ &  N/A &  $\boldsymbol{87.01\pm0.20}$ &  N/A \\
    \midrule
    OCR (baseline)$^{\dagger}$ & \multirow{5}[0]{*}{HR$48$} &  $80.70$ &   &  $88.11$ &  \\
     OCR (baseline) &  &  $80.18\pm0.12$ &  * &  $87.36\pm0.30$ &  * \\
     OCR-SE ($2018$) &       &  $80.32\pm0.20$ &  * &  $\underline{\textit{87.66}}\pm\underline{\textit{0.14}}$ &  * \\
     OCR-ECA ($2020$) &       &  $\underline{\textit{80.35}}\pm\underline{\textit{0.15}}$ &  * &  $87.60\pm0.26$ &  * \\
     OCR-GPCA (ours) &       &  $\boldsymbol{80.65\pm0.13}$ &  N/A &  $\boldsymbol{88.07\pm0.32}$ &  N/A \\
    \bottomrule
    \end{tabular}
  \label{tab:cityscapes}
  \vspace{-4mm}
\end{table}

\vspace{-4mm}
\subsection{Dataset}\label{ssec:segdataset}

{We utilize the CityScapes~\cite{cordts2016cityscapes} dataset for evaluations. It contains $19$ classes with $3,475$ annotated images for training and validating, and $1,525$ images for test.}

\vspace{-4mm}
\subsection{Implementation Details}\label{ssec:cityscapesim}

{On the CityScapes dataset, the object-contextual representation network (OCRNet)~\cite{yuan2019object} model, which applies HRNetV$2$-W$18$ and HRNetV$2$-W$48$~\cite{sun2019deep} models as its backbones, was used as the base model for the attention modules. Meanwhile, all the attention modules were only applied after the OCR module for enhancing feature representation. During the training process, we followed the original strategy in MMSegmentation~\cite{mmseg2020} {with $80,000$ iterations and batch size of $4$ due to limited GPU resources ($8$ for the original setting)}. The two backbones with all the attention modules were conducted three times with random initialization, respectively. The means and the standard deviations of the mean intersection over union (mIoU, \%) and mean accuracy (mACC, \%) are reported.}

\vspace{-4mm}
\subsection{Performance on CityScapes Dataset}\label{ssec:cityscapes}

{The experimental results of the CityScapes dataset are listed in Table~\ref{tab:cityscapes}. According to Table~\ref{tab:cityscapes}, the proposed GPCA module can achieve $80.09\%$ and $87.01\%$ for mIoU and mAcc with HR$18$, respectively. With HR$48$, the mIoU and mACC are $80.65\%$ and $88.07\%$, respectively. All these values are larger than those obtained by the referred methods. The $p$-values of the student's-$t$ test between the proposed GPCA module and the reimplemented  indicate the statistically significant improvement on segmentation performance.}

{It is worth to note that our mean mIoU and mAcc with the HRNetV$2$-W$48$ are both slightly lower than the ones reported in~\cite{mmseg2020}, which may be caused by the smaller batch size in our implementations. However, there is no statistically significant difference between them. Furthermore, we find the highest mIoU and mAcc of ours out of three times of simulations are $80.81\%$ and $88.43\%$, respectively. These results are higher than the officially reported ones. This indicates that, although the proposed GPCA module was affected by the smaller batch size setting, it still has competitive performance compared with the official results.}

\begin{figure*}[!t]
    \centering
    \begin{subfigure}[t]{0.46\linewidth}
        \centering
        \includegraphics[width=1\linewidth]{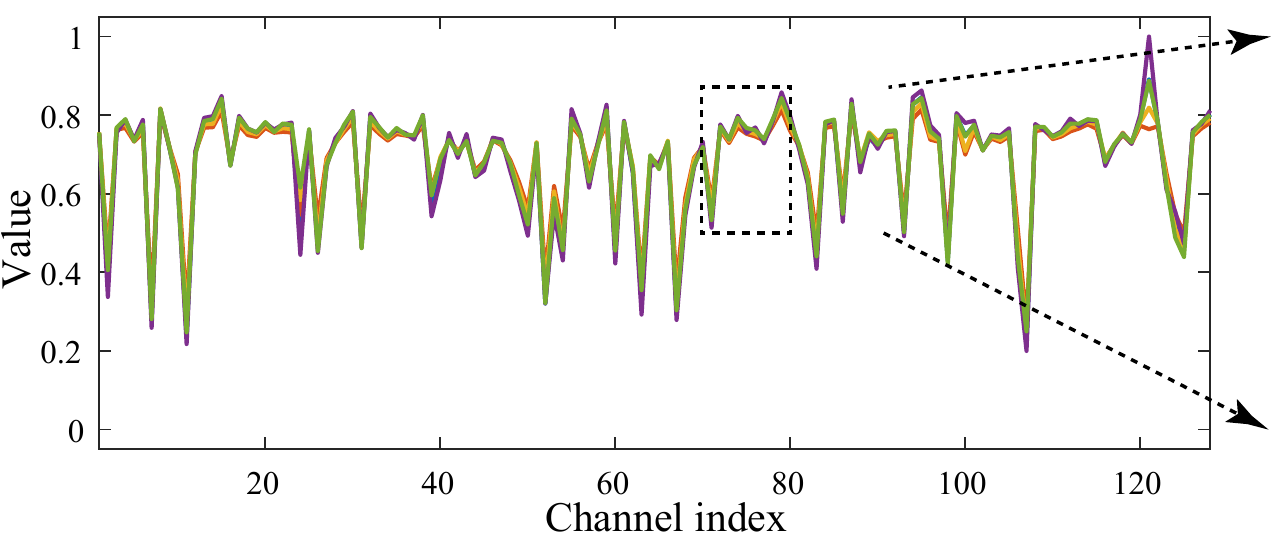}
        \vspace{-6mm}
    \subcaption{Attention mask values of SENet$\_3\_1$}
    \end{subfigure}
    \begin{subfigure}[t]{0.26\linewidth}
        \centering
        \includegraphics[width=1\linewidth]{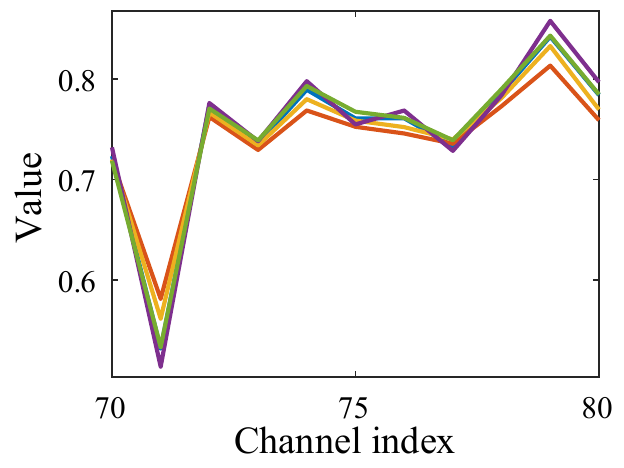}
        \vspace{-6mm}
    \subcaption{Channel $70$-$80$ of SENet$\_3\_1$}
    \end{subfigure}
    \begin{subfigure}[t]{0.26\linewidth}
        \centering
        \includegraphics[width=1\linewidth]{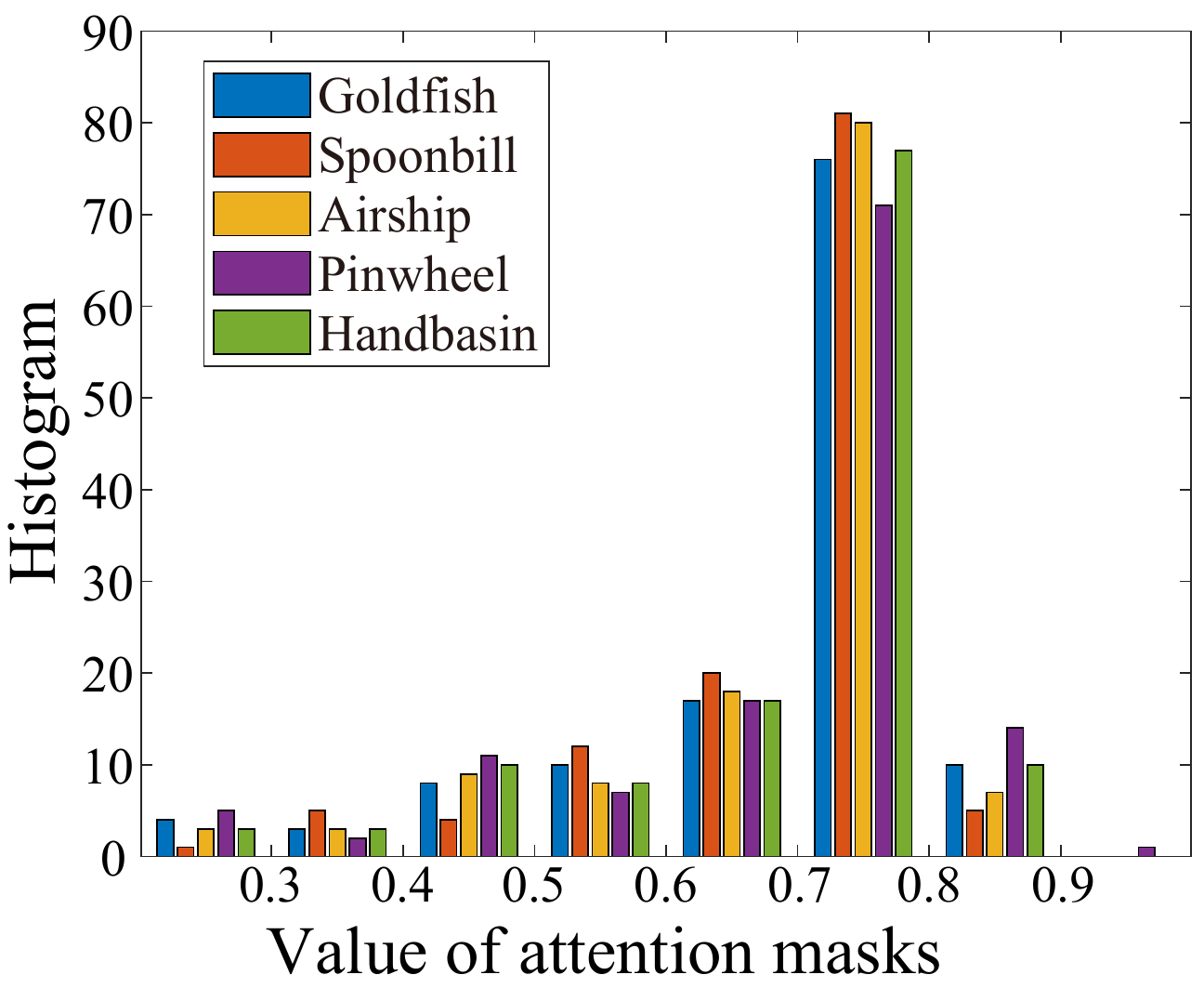}
        \vspace{-6mm}
    \subcaption{Histogram of SENet$\_3\_1$}
    \end{subfigure}
    \begin{subfigure}[t]{0.46\linewidth}
        \centering
        \includegraphics[width=1\linewidth]{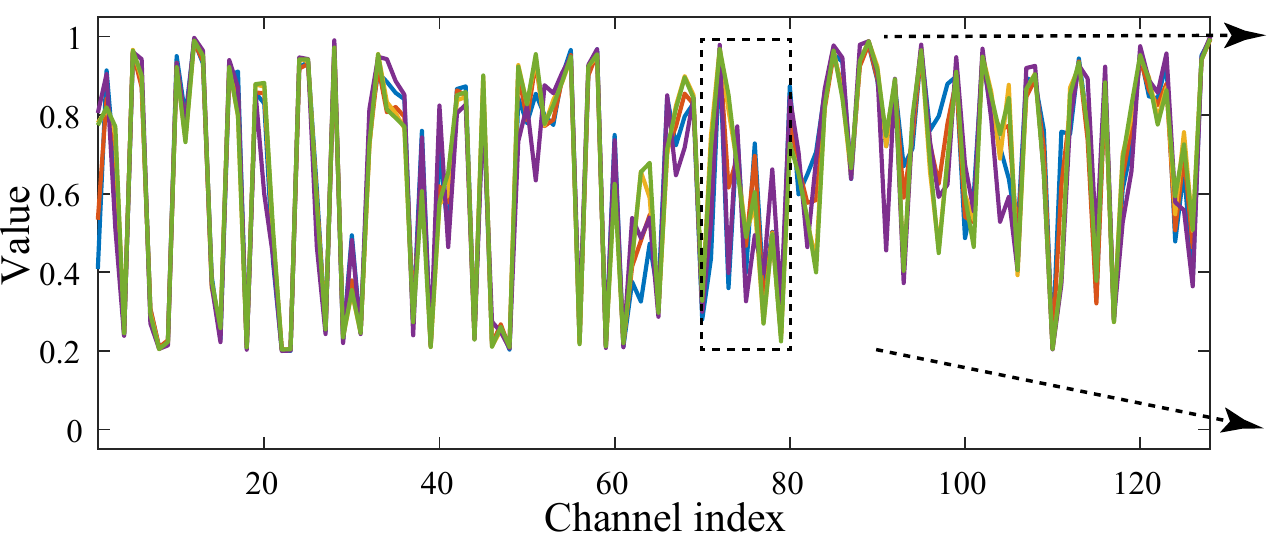}
        \vspace{-6mm}
    \subcaption{Attention mask values of GPCA$\_3\_1$}
    \end{subfigure}
    \begin{subfigure}[t]{0.26\linewidth}
        \centering
        \includegraphics[width=1\linewidth]{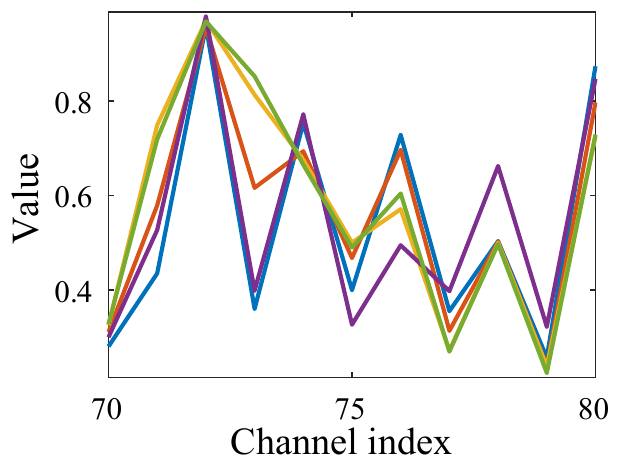}
        \vspace{-6mm}
    \subcaption{Channel $70$-$80$ of GPCA$\_3\_1$}
    \end{subfigure}
    \begin{subfigure}[t]{0.26\linewidth}
        \centering
        \includegraphics[width=1\linewidth]{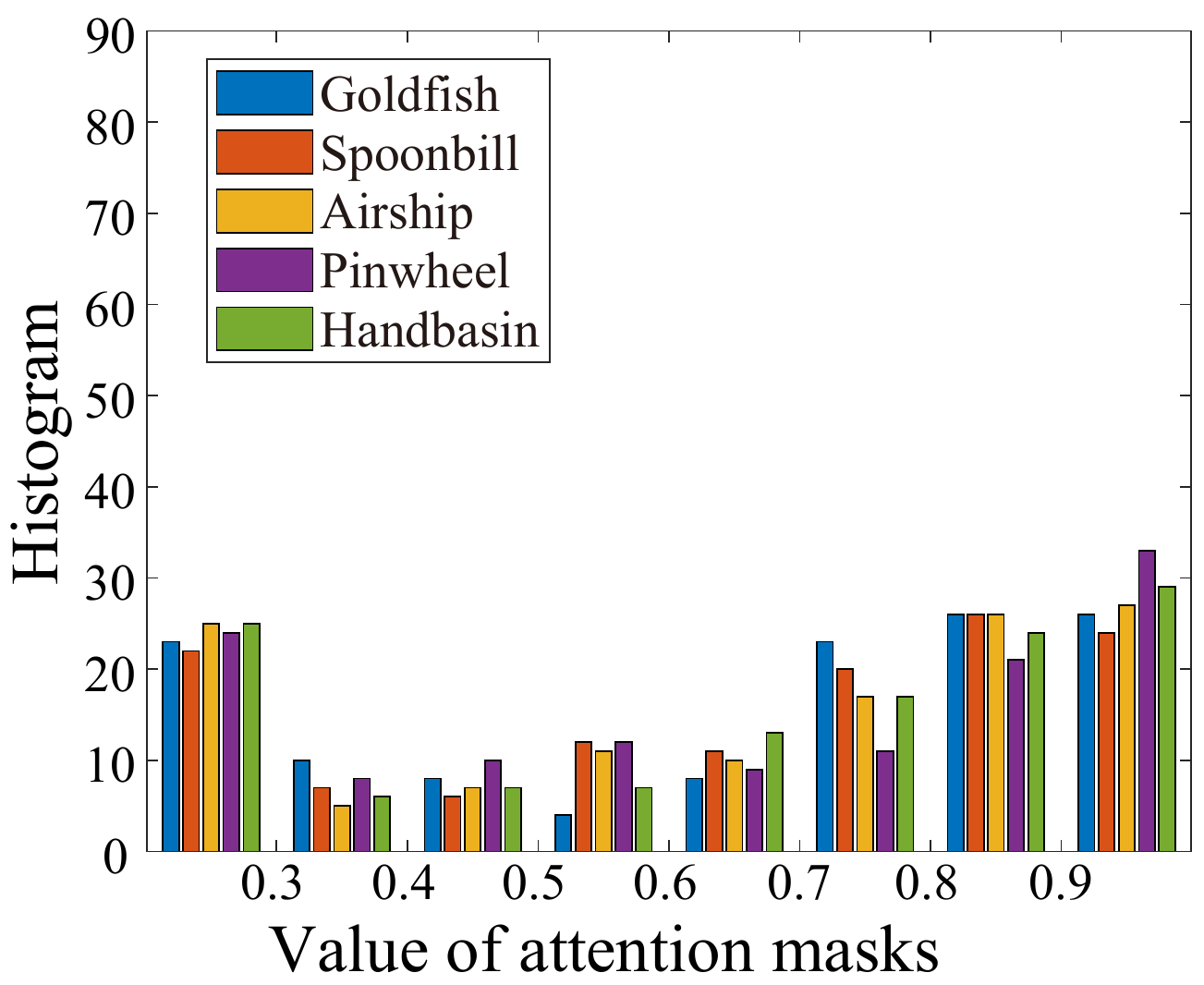}
        \vspace{-6mm}
        \subcaption{Histogram of GPCA$\_3\_1$}
    \end{subfigure}
    \vspace{-2mm}
    \caption{The values of the attention masks in (a) the SENet and (d) the GPCA modules with the ResNet$18$ on the ImageNet-$32\!\times\!32$ dataset. We also zoomed in channel $70$-$80$ (dashed boxes in (a) and (d)) for better illustration in (b) and (e), respectively. The histograms of the values are illustrated in (c) for the SENet and in (f) for the GPCA, respectively. The modules in each subfigure is named as ``attention name\_stageID\_blockID''.}\label{fig:gpcamasks}
    \vspace{-4mm}
\end{figure*}

\begin{figure*}[!t]
    \centering
    \begin{subfigure}[t]{0.85\linewidth}
        \centering
        \includegraphics[width=1\linewidth]{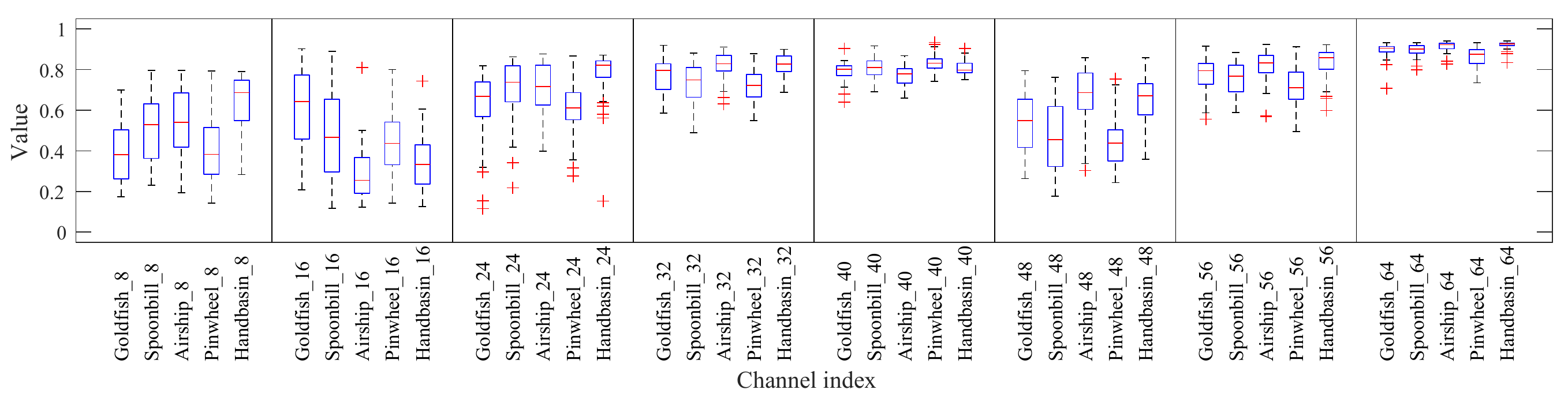}
        \vspace{-6mm}
        \subcaption{GPCA$\_2\_1$}
    \end{subfigure}
    \begin{subfigure}[t]{0.85\linewidth}
        \centering
        \includegraphics[width=1\linewidth]{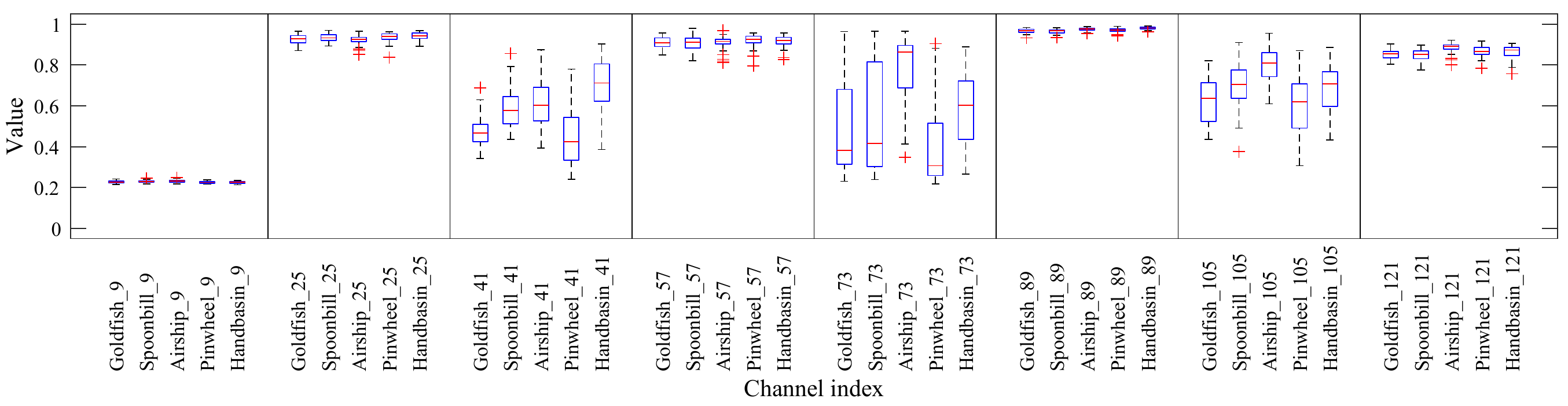}
        \vspace{-6mm}
        \subcaption{GPCA$\_3\_1$}
    \end{subfigure}
    \vspace{-3mm}
    \caption{Boxplots of the attention masks at different depths in the GPCA module with the ResNet$18$ on the ImageNet-$32\times32$ dataset. The \emph{x}-axis denotes the channel index. The modules in each subfigure are named as GPCA$\_$stageID$\_$blockID. GPCA$\_2\_1$ and GPCA$\_3\_1$ are selected as examples. For better illustration, we selected parts of the channels and draw the corresponding boxplots.}\label{fig:gpcamaskboxplots}
    \vspace{-6mm}
\end{figure*}

\vspace{-4mm}
\section{Analysis of the GPCA Module}\label{sec:analysis}

\subsection{The Role of the GPCA Module}\label{ssec:role}

To investigate the role of the GPCA in CNNs, we illustrate the distribution of the values of the attention masks learned by {the SENet} and the proposed GPCA module for different classes. We selected five classes,~\emph{i.e.}, goldfish, spoonbill, airship, pinwheel, and washbasin, from the ImageNet-$32\!\times\!32$ dataset and illustrate the averaged values of the attention masks on each channel of all the test samples {and the histograms of the values} in the corresponding classes in Figure~\ref{fig:gpcamasks}. ResNet$18$ has been selected as the base model. The modules in each subfigure are named as GPCA$\_$stageID$\_$blockID. {Note that the same channels between the SENet and the proposed GPCA do not always indicate same features, which means we cannot directly compare the values of their attention masks channel by channel.} {In addition to the averaged values of the attention masks, we select two blocks, \emph{i.e.}, GPCA$\_2\_1$ and GPCA$\_3\_1$, to show their boxplots in Figure~\ref{fig:gpcamaskboxplots}. For better illustration, eight channels of the aforementioned blocks are selected at equal intervals.}

{It can be observed that, in most of the channels, the masks of GPCA for different classes are more distinct than those of SENet. For different channels, the GPCA masks vary dramatically along the channel index. However, for more than half of the channels, the SENet masks are similar in values. These observations suggest that the proposed GPCA module has better ability in learning the channel information than the SENet. This is because different values of the attention masks (\emph{i.e.}, importance weights) are assigned to each channel by GPCA and show their distinct roles, while the SENet tends to assign the channels equal weights and, therefore, the attention scheme does not work well.}

Thus, the effectiveness of the GPCA module can be clarified since the GPCA modules perform as channel-wise feature selectors and play distinct and important roles in different depths of the ResNet$18$ model. In addition, similar phenomena can be also observed on the other datasets, and in the VGG$16$ and the ResNet$34$ models, respectively. {We should note that the attention masks, which perform as weights of channels, reflect the importance of the corresponding channels, rather than directly demonstrating the correlations between the channels, although the attention masks are estimated according to the channel correlations.}

\vspace{-4mm}
\subsection{Ability of Learning the Channel Correlations}\label{ssec:nll}

{In this section, we intuitively evaluate the attention modules’ abilities in learning the channel correlations to clarify the actual benefits and advantages of the proposed GPCA module. Given that the final FC layers are replaced by convolutional layers following a global average pooling (GAP) and the output number of channels of the last convolutional layer is aligned to the number of classes in the WSOL task (see Section~\ref{ssec:locimplement}), the attention modules can be employed before the GAP and play a role in selecting a true channel (if possible) for the prediction by setting higher attention masks for the ground truth classes without supervision. In this case, the classification labels can be treated as the pseudo targets of the attention modules and evaluate the attention modules explicitly.} 

{We conduct a group of experiments with standard negative log-likelihood (NLL) metric~\cite{maddox2019a} {(the smaller the better)}. According to Table~\ref{tab:nll}, the GPCA module obtains the smallest NLL values on all the three base models, compared with the SENet and the SRP. This indicates that the GPCA can learn the channel correlations better than the referred methods. It clearly explains why the GPCA essentially performs better than the full connection-based method (\emph{e.g.}, the SENet) and the convolution- and pooling-based method (\emph{e.g.}, the SRP).}

\begin{table}[!t]
\vspace{1mm}
  \centering
  \caption{Analysis of GPCA module by comparing the NLLs on the CUB-$200$-$2011$ dataset. The means and standard deviations of three rounds of evaluations are listed. The best results for each base model are marked in \textbf{bold}.}
  \vspace{-2mm}
  \resizebox{\linewidth}{!}{
    \begin{tabular}{@{}l@{}c@{}c@{}c@{}}
    \toprule
    \multicolumn{1}{c}{Method} &  VGG$16$ &  MobileNet &  ResNet$50$ \\
    \midrule
     SENet ($2018$) &  $4.6127\pm0.4220$ &  $3.6650\pm0.0899$ &  $3.6200\pm0.1052$ \\
     SRP ($2019$) &  $5.0990\pm0.0016$ &  $5.0961\pm0.0052$ &  $5.0814\pm0.0179$ \\
     GPCA (ours) & \ \  $\boldsymbol{3.5969\pm0.2217}$ &  \ \ $\boldsymbol{3.5537\pm0.0555}$ &  \ \ $\boldsymbol{3.4729\pm0.2390}$ \\
    \bottomrule
    \end{tabular}}
  \label{tab:nll}
  \vspace{-4mm}
\end{table}

\begin{table*}[!t]
    \centering
    \caption{Comparisons of parameter numbers (\#Params.), computational complexities, {and training and inference times per epoch (second/epoch)}. $C$ is number of channels, $H$ and $W$ are height and width of feature maps, and $\left|\cdot\right|_{\text{ood}}$ indicates the nearest odd number. {The training and inference times were evaluated in the ResNet$50$ on the Cifar-$100$ dataset with batch size as $256$ using one NVIDIA GTX$1080$Ti.}}
    \vspace{-2mm}
    \begin{tabular}{ccccc}
        \toprule
         Model &  \#Params. &  Computational complexity &  Training time &  Inference time\\ 
        \midrule
         SENet ($2018$) &  $\frac{2\times C^2}{r}$ with $r=4$ &  $O(C^2)$ &  $26.41$ &  $1.49$ \\
         NLNet ($2018$) &  $2C^2$ &  $O(C^2+H^2W^2C)$ &  $44.38$ &  $4.52$\\
         BAM ($2018$) &  $\frac{3\times C^2+C}{r}+\frac{9C^2}{r^2}$ with $r=4$ &  $O(C^2)$ &  $21.16$ &  $2.13$\\
         CBAM ($2018$) &  $\frac{2C^2}{r}+98$ with $r=4$ &  $O(C^2)$ &  $36.63$ &  $3.64$\\
         ECA ($2020$) &  $\left|\frac{\log_2 C}{\gamma}+\frac{b}{\gamma}\right|_{\text{ood}}$ with $\gamma=2,b=1$ &  $O(C\ln C)$ &  $22.30$ &  $2.43$\\
         GCT ($2020$) &  $3\times C$ &  $O(C^2)$ &  $25.53$ &  $2.59$\\
         GPCA (ours) &  $4$ &  $O(C^3)$ &  $53.19$ &  $5.22$\\
        \bottomrule
    \end{tabular}
    \label{tab:complexity}
    \vspace{-2mm}
\end{table*}

\begin{table*}[!t]
  \centering
  \caption{Two simplified versions of the proposed GPCA module by local Gaussian process (Local) and multihead attention (MHA). We compare their performance on three datasets, including Cifar-$10$/-$100$ and ImageNet. Moreover, number of parameters, computational complexity, and training and inference time per epoch (second/epoch) are also compared. $h$ is head number in the MHA.}
  \vspace{-2mm}
  \scriptsize
    \begin{tabular}{lcccccccc}
    \toprule
    \multicolumn{1}{c}{\multirow{2}[0]{*}{Method}} & \multirow{2}[0]{*}{Cifar-$10$} & \multirow{2}[0]{*}{Cifar-$100$} & \multicolumn{2}{c}{ImageNet} & \multirow{2}[0]{*}{\#Params.} & {Computational} & {Training} & {Inference} \\
      & & &  Top-$1$ &  Top-$5$ & &  complexity &  time &  time \\
    \midrule
     GPCA-Local & $93.73\pm0.21$ & $72.02\pm0.15$ & $76.36\pm0.13$ & $93.41\pm0.17$ & $4$ & $O((\ln C)^3)$ & $24.15$ & $2.39$ \\
     GPCA-MHA & $94.11\pm0.09$ & $72.37\pm0.12$ & $76.42\pm0.16$ & $93.38\pm0.26$ & $4$ & $O((\frac{C}{h})^3)$ & $33.95$ & $3.30$ \\
     GPCA & $94.93\pm0.01$ & $73.97\pm0.02$ & $77.57\pm0.07$ & $93.97\pm0.06$ & $4$ & $O(C^3)$ & $53.19$ & $5.22$ \\
    \bottomrule
    \end{tabular}
  \label{tab:lgp}
  \vspace{-4mm}
\end{table*}

\vspace{-4mm}
\subsection{{Comparisons of Parameter Numbers and Computational Complexities}}\label{ssec:paramcomputation}

{We compare the numbers of parameters and computational complexity of the proposed GPCA module and the referred attention modules, respectively, as shown in Table~\ref{tab:complexity}. The proposed GPCA module only introduces the four parameters in $\boldsymbol{\Theta}$, which is much less than the other attention modules, and the computational complexity of it is $O(C^3)$, which is a little higher than them. {We further compare the training and the inference times per epoch (in second/epoch). It can be observed that the training and inference times of GPCA are not increased too much and have similar values to those of the NLNet.}}

\vspace{-4mm}
\subsection{Two Simplified Versions of GPCA}\label{ssec:lgp}

{In order to evaluate the trade-off between the performance and the computational complexity, we propose two simplified versions of the GPCA with lower computational complexities. The first version introduces the local GP as the prior, in which the GP prior is conducted in a near neighbourhood of the channels, namely GPCA-Local. In the GPCA-Local, $\left|\frac{\log_2 C}{\gamma}+\frac{b}{\gamma}\right|_{\text{ood}}$ channels with $\gamma=2,b=1$ were selected for one channel as the local neighbourhood, similar to the ECA module. Another version introduces the multihead attention (MHA)-like framework, assigning channels into different groups and conducting the GPCA inside each group (denoted as GPCA-MHA from hereby). In the GPCA-MHA, we assigned $16$ consecutive channels into one group for each block and obtained $\frac{C}{16}$ groups in total. The comparisons between these two versions and the full GPCA are shown in Table~\ref{tab:lgp}. Although these two simplified versions can significantly reduce the computational complexities and the time cost, the performance in the image classification task is also decreased. This is due to a trade-off between the performance and the computational complexity.}

\vspace{-4mm}
\section{Conclusions}\label{sec:conclusions}

In this paper, we proposed a Gaussian process embedded channel attention (GPCA) module to learn the correlations among the channels in CNNs via a probabilistic way. In the proposed GPCA, the output channel attention masks were assumed as beta distributed variables. We adapted a Sigmoid-Gaussian approximation, in which the Gaussian distributed variables were transferred and bounded into the interval of $[0,1]$ by a Sigmoid function. The Gaussian process was utilized to model the correlations among different channels and applied as a prior of the channel attention masks. In this case, the proposed GPCA module can be intuitively and reasonably clarified in a probabilistic way. Although the beta distribution has difficulty to be integrated into the end-to-end training of the CNNs, we utilized an appropriate approximation of the beta distribution to derive an analytically tractable solution to facilitate the calculations. 

Furthermore, experimental results show that the proposed GPCA can statistically significantly improve the classification accuracies with three widely used base models on five benchmark image classification datasets, compared with eight referred attention modules. In addition, statistically significant improvements have also been obtained in WSOL, object detection, {and semantic segmentation} tasks, compared with recently proposed state-of-the-art methods.

\ifCLASSOPTIONcompsoc
  \section*{Acknowledgments}
\else
  \section*{Acknowledgment}
\fi

This work was supported in part by the National Key R\&D Program of China under Grant $2019$YFF$0303300$, Subject II No. $2019$YFF$0303302$, and $2020$AAA$0105200$, in part by National Natural Science Foundation of China (NSFC) under Grant $61922015$, $61773071$, U$19$B$2036$, in part by Beijing Natural Science Foundation Project under Grant Z$200002$, and in part by BUPT Excellent Ph.D. Students Foundation No. CX$2020105$.

\ifCLASSOPTIONcaptionsoff
  \newpage
\fi

\bibliographystyle{IEEEbib}
\footnotesize
\bibliography{GPCA_IEEE_trans}

\begin{thebibliography}{10}

\bibitem{simonyan2015very}
K.~Simonyan and A.~Zisserman,
\newblock ``Very deep convolutional networks for large-scale image
  recognition,''
\newblock in {\em International Conference on Learning Representations}, 2015.

\bibitem{he2016deep}
K.~He, X.~Zhang, S.~Ren, and J.~Sun,
\newblock ``Deep residual learning for image recognition,''
\newblock in {\em Computer Vision and Pattern Recognition}, 2016, pp. 770--778.

\bibitem{huang2017densely}
G.~Huang, Z.~Liu, L.~V. Der~Maaten, and K.~Q. Weinberger,
\newblock ``Densely connected convolutional networks,''
\newblock in {\em Computer Vision and Pattern Recognition}, 2017, pp.
  2261--2269.

\bibitem{peng2019few}
Z.~Peng, Z.~Li, J.~Zhang, Y.~Li, G.-J. Qi, and J.~Tang,
\newblock ``Few-shot image recognition with knowledge transfer,''
\newblock in {\em International Conference on Computer Vision (ICCV)}, 2019,
  pp. 441--449.

\bibitem{liu2016ssd}
W.~Liu, D.~Anguelov, D.~Erhan, C.~Szegedy, S.~E. Reed, C.~Fu, and A.~C. Berg,
\newblock ``{SSD}: {S}ingle shot multibox detector,''
\newblock in {\em European Conference on Computer Vision}, 2016, pp. 21--37.

\bibitem{chen2018deeplab}
L.~Chen, G.~Papandreou, I.~Kokkinos, K.~Murphy, and A.~L. Yuille,
\newblock ``{DeepLab}: {S}emantic image segmentation with deep convolutional
  nets, atrous convolution, and fully connected {CRFs},''
\newblock {\em IEEE Transactions on Pattern Analysis and Machine Intelligence},
  vol. 40, no. 4, pp. 834--848, 2018.

\bibitem{yuan2019object}
Y.~Yuan, X.~Chen, and J.~Wang,
\newblock ``Object-contextual representations for semantic segmentation,''
\newblock {\em arXiv preprint arXiv:1909.11065}, 2019.

\bibitem{li2019deep}
Z.~Li, J.~Tang, and T.~Mei,
\newblock ``Deep collaborative embedding for social image understanding,''
\newblock {\em IEEE Transactions on Pattern Analysis and Machine Intelligence},
  vol. 41, no. 9, pp. 2070--2083, 2019.

\bibitem{hinton12}
G.~E. Hinton, N.~Srivastava, A.~Krizhevsky, I.~Sutskever, and R.~Salakhutdinov,
\newblock ``Improving neural networks by preventing co-adaptation of feature
  detectors,''
\newblock {\em arXiv}, 2012.

\bibitem{vaswani2017attention}
A.~Vaswani, N.~Shazeer, N.~Parmar, J.~Uszko-reit, L.~Jones, A.~N. Gomez, Ł.
  Kaiser, and I.~Polosukhin,
\newblock ``Attention is all you need,''
\newblock in {\em Advances in Neural Information Processing Systems}, 2017, pp.
  5998--6008.

\bibitem{hu2018squeeze}
J.~Hu, L.~Q. Shen, and G.~Sun,
\newblock ``Squeeze-and-excitation networks,''
\newblock in {\em Computer Vision and Pattern Recognition}, 2018, pp.
  7132--7141.

\bibitem{jaderberg2015spatial}
M.~Jaderberg, K.~Simonyan, A.~Zisserman, and K.~Kavukcuoglu,
\newblock ``Spatial transformer networks,''
\newblock in {\em Neural Information Processing Systems}, 2015, pp. 2017--2025.

\bibitem{woo2018cbam}
S.~Woo, J.~Park, J.-Y. Lee, and I.~S. Kweon,
\newblock ``{CBAM}: {C}onvolutional block attention module,''
\newblock in {\em European Conference on Computer Vision}, 2018, pp. 3--19.

\bibitem{hu2018gather-excite}
J.~Hu, L.~Shen, S.~Albanie, G.~Sun, and A.~Vedaldi,
\newblock ``Gather-{E}xcite: {E}xploiting feature context in convolutional
  neural networks,''
\newblock in {\em Neural Information Processing Systems}, 2018, pp. 9401--9411.

\bibitem{sun2018multi-attention}
M.~Sun, Y.~Yuan, F.~Zhou, and E.~Ding,
\newblock ``Multi-attention multi-class constraint for fine-grained image
  recognition,''
\newblock in {\em European Conference on Computer Vision}, 2018, pp. 834--850.

\bibitem{park2018bam}
J.~Park, S.~Woo, J.~Lee, and I.~S. Kweon,
\newblock ``{BAM}: {B}ottleneck attention module.,''
\newblock in {\em British Machine Vision Conference}, 2018, p. 147.

\bibitem{huang2019dianet}
Z.~Huang, S.~Liang, M.~Liang, and H.~Yang,
\newblock ``{DIANet}: {D}ense-and-implicit attention network.,''
\newblock {\em arXiv}, 2019.

\bibitem{yang2019two-level}
Y.~Yang, X.~Wang, Q.~Zhao, and T.~Sui,
\newblock ``Two-level attentions and grouping attention convolutional network
  for fine-grained image classification,''
\newblock {\em Applied Sciences}, vol. 9, no. 9, pp. 1939, 2019.

\bibitem{luo2019stochastic}
M.~Luo, G.~Wen, Y.~Hu, D.~Dai, and Y.~Xu,
\newblock ``Stochastic region pooling: {M}ake attention more expressive.,''
\newblock {\em arXiv}, 2019.

\bibitem{lopez2019pay}
P.~R. Lopez, D.~V. Dorta, G.~C. Preixens, J.~M.~G. Sitjes, F.~X.~R. Marva, and
  J.~Gonzalez,
\newblock ``Pay attention to the activations: a modular attention mechanism for
  fine-grained image recognition,''
\newblock {\em IEEE Transactions on Multimedia}, 2019.

\bibitem{bishop06}
C.~M. Bishop,
\newblock {\em Pattern recognition and machine learning},
\newblock Springer Science+Business Media LLC., 2006.

\bibitem{Rasmussen2006gaussian}
C.~E. Rasmussen and C.~K.~I. Williams,
\newblock {\em Gaussian processes for machine learning},
\newblock The MIT Press, 2006.

\bibitem{chen2019hybrid}
B.~Chen and W.~Deng,
\newblock ``Hybrid-attention based decoupled metric learning for zero-shot
  image retrieval,''
\newblock in {\em Computer Vision and Pattern Recognition}, 2019, pp.
  2745--2754.

\bibitem{fang2019bilinear}
P.~Fang, J.~Zhou, S.~Roy, L.~Petersson, and M.~Harandi,
\newblock ``Bilinear attention networks for person retrieval,''
\newblock in {\em International Conference on Computer Vision}, 2019, pp.
  8029--8038.

\bibitem{zhao2019pyramid}
T.~Zhao and X.~Wu,
\newblock ``Pyramid feature attention network for saliency detection,''
\newblock in {\em Computer Vision and Pattern Recognition}, 2019, pp.
  3080--3089.

\bibitem{chen2018reverse}
S.~Chen, X.~Tan, B.~Wang, and X.~Hu,
\newblock ``Reverse attention for salient object detection,''
\newblock in {\em European Conference on Computer Vision}, 2018, pp. 236--252.

\bibitem{linsley2018global}
D.~Linsley, D.~Shiebler, S.~Eberhardt, and T.~Serre,
\newblock ``Global-and-local attention networks for visual recognition,''
\newblock {\em arXiv}, 2018.

\bibitem{chen2019mixed}
B.~Chen, W.~Deng, and J.~Hu,
\newblock ``Mixed high-order attention network for person re-identification,''
\newblock in {\em International Conference on Computer Vision}, 2019, pp.
  371--381.

\bibitem{chen2019self}
G.~Chen, C.~Lin, L.~Ren, J.~Lu, and J.~Zhou,
\newblock ``Self-critical attention learning for person re-identification,''
\newblock in {\em International Conference on Computer Vision}, 2019, pp.
  9636--9645.

\bibitem{yan2018multi-level}
Y.~Yan, B.~Ni, J.~Liu, and X.~Yang,
\newblock ``Multi-level attention model for person re-identification,''
\newblock {\em Pattern Recognition Letters}, 2018.

\bibitem{zhu2018end-to-end}
Z.~Zhu, W.~Wu, W.~Zou, and J.~Yan,
\newblock ``End-to-end flow correlation tracking with spatial-temporal
  attention,''
\newblock in {\em Computer Vision and Pattern Recognition}, 2018, pp. 548--557.

\bibitem{muqeet2019hybrid}
A.~Muqeet, T.~B. Iqbal, and S.~Bae,
\newblock ``Hybrid residual attention network for single image super
  resolution,''
\newblock {\em arXiv}, 2019.

\bibitem{jang2019densenet}
D.~Jang and R.~Park,
\newblock ``{DenseNet} with deep residual channel-attention blocks for single
  image super resolution,''
\newblock in {\em Computer Vision and Pattern Recognition}, 2019.

\bibitem{dai2019second}
T.~Dai, J.~Cai, Y.~Zhang, S.~Xia, and L.~Zhang,
\newblock ``Second-order attention network for single image super-resolution,''
\newblock in {\em Computer Vision and Pattern Recognition}, 2019, pp.
  11057--11066.

\bibitem{fu2019dual}
J.~Fu, J.~Liu, H.~Tian, Y.~Li, Y.~Bao, Z.~Fang, and H.~Lu,
\newblock ``Dual attention network for scene segmentation,''
\newblock in {\em Computer Vision and Pattern Recognition}, 2019, pp.
  3141--3149.

\bibitem{huang2019ccnet}
Z.~Huang, X.~Wang, L.~Huang, C.~Huang, Y.~Wei, and W.~Liu,
\newblock ``{CCNet}: {C}riss-cross attention for semantic segmentation,''
\newblock in {\em International Conference on Computer Vision}, 2019, pp.
  603--612.

\bibitem{lu2019see}
X.~Lu, W.~Wang, C.~Ma, J.~Shen, L.~Shao, and F.~Porikli,
\newblock ``See more, know more: {U}nsupervised video object segmentation with
  co-attention siamese networks,''
\newblock in {\em Computer Vision and Pattern Recognition}, 2019, pp.
  3618--3627.

\bibitem{bello2019attention}
I.~Bello, B.~Zoph, Q.~Le, A.~Vaswani, and J.~Shlens,
\newblock ``Attention augmented convolutional networks,''
\newblock in {\em International Conference on Computer Vision}, 2019, pp.
  3285--3294.

\bibitem{cao2019gcnet}
Y.~Cao, J.~Xu, S.~Lin, F.~Wei, and H.~Hu,
\newblock ``{GCNet}: {N}on-local networks meet squeeze-excitation networks and
  beyond,''
\newblock in {\em International Conference on Computer Vision Workshop}, 2019,
  pp. 1971--1980.

\bibitem{wang2020ecanet}
Q.~Wang, B.~Wu, P.~Zhu, P.~Li, W.~Zuo, and Q.~Hu,
\newblock ``{ECA-Net}: {E}fficient channel attention for deep convolutional
  neural networks,''
\newblock in {\em Computer Vision and Pattern Recognition}, 2020.

\bibitem{yang2020gated}
Z.~Yang, L.~Zhu, Y.~Wu, and Y.~Yang,
\newblock ``Gated channel transformation for visual recognition,''
\newblock in {\em Computer Vision and Pattern Recognition}, 2020.

\bibitem{hochreiter1997long}
S.~Hochreiter and J.~Schmidhuber,
\newblock ``Long short-term memory,''
\newblock {\em Neural Computation}, vol. 9, no. 8, pp. 1735--1780, 1997.

\bibitem{sudhakaran2019lsta}
S.~Sudhakaran, S.~Escalera, and O.~Lanz,
\newblock ``{LSTA}: {L}ong short-term attention for egocentric action
  recognition,''
\newblock in {\em Computer Vision and Pattern Recognition}, 2019, pp.
  9946--9955.

\bibitem{li2019selective}
X.~Li, W.~Wang, X.~Hu, and J.~Yang,
\newblock ``Selective kernel networks,''
\newblock in {\em Computer Vision and Pattern Recognition}, 2019.

\bibitem{chen2020dynamic}
Y.~Chen, X.~Dai, M.~Liu, D.~Chen, L.~Yuan, and Z.~Liu,
\newblock ``Dynamic convolution: {A}ttention over convolution kernels,''
\newblock in {\em Computer Vision and Pattern Recognition}, 2020.

\bibitem{lin2015bilinear}
T.~Lin, A.~Roychowdhury, and S.~Maji,
\newblock ``Bilinear {CNN} models for fine-grained visual recognition,''
\newblock in {\em International Conference on Computer Vision}, 2015, pp.
  1449--1457.

\bibitem{zheng2020blobal}
Z.~Zheng, G.~An, D.~Wu, and Q.~Ruan,
\newblock ``Global and local knowledge-aware attention network for action
  recognition,''
\newblock {\em IEEE Transactions on Neural Networks and Learning Systems}, pp.
  1--14, 2020.

\bibitem{yang2019attention}
L.~Yang, Q.~Song, Y.~Wu, and M.~Hu,
\newblock ``Attention inspiring receptive-fields network for learning invariant
  representations,''
\newblock {\em IEEE Transactions on Neural Networks and Learning Systems}, vol.
  30, no. 6, pp. 1744--1755, 2019.

\bibitem{wang2018non}
X.~Wang, R.~Girshick, A.~Gupta, and K.~He,
\newblock ``Non-local neural networks,''
\newblock in {\em Computer Vision and Pattern Recognition}, June 2018, pp.
  7794--7803.

\bibitem{cheo2019attention}
J.~Choe and H.~Shim,
\newblock ``Attention-based dropout layer for weakly supervised object
  localization,''
\newblock in {\em Computer Vision and Pattern Recognition}, 2019, pp.
  2214--2223.

\bibitem{choe2020attention}
J.~Choe, S.~Lee, and H.~Shim,
\newblock ``Attention-based dropout layer for weakly supervised single object
  localization and semantic segmentation,''
\newblock {\em IEEE Transactions on Pattern Analysis and Machine Intelligence},
  2020.

\bibitem{zhao2020exploring}
H.~Zhao, J.~Jia, and V.~Koltun,
\newblock ``Exploring self-attention for image recognition,''
\newblock in {\em Computer Vision and Pattern Recognition}, 2020.

\bibitem{xie2019soft}
J.~Xie, Z.~Ma, G.~Zhang, J.-H. Xue, Z.-H. Tan, and J.~Guo,
\newblock ``Soft dropout and its variational {B}ayes approximation,''
\newblock in {\em IEEE International Workshop on Machine Learning for Signal
  Processing}, 2019.

\bibitem{maddison2017the}
C.~J. Maddison, A.~Mnih, and Y.~W. Teh,
\newblock ``The concrete distribution: {A} continuous relaxation of discrete
  random variables,''
\newblock {\em ArXiv}, 2017.

\bibitem{mackey1992the}
D.~J.~C. MacKay,
\newblock ``The evidence framework applied to classification networks,''
\newblock {\em Neural Computation}, vol. 4, no. 5, pp. 720--736, 1992.

\bibitem{krizhevsky09cifar}
A.~Krizhevsky,
\newblock ``Learning multiple layers of features from tiny images,''
\newblock techreport, CIFAR, 2009.

\bibitem{vinyals2016matching}
O.~Vinyals, C.~Blundell, T.~P. Lillicrap, K.~Kavukcuoglu, and D.~Wierstra,
\newblock ``Matching networks for one shot learning,''
\newblock {\em arXiv}, 2016.

\bibitem{denoord2016pixel}
A.~V. Den~Oord, N.~Kalchbrenner, and K.~Kavukcuoglu,
\newblock ``Pixel recurrent neural networks,''
\newblock {\em arXiv}, 2016.

\bibitem{russakovsky2015imagenet}
O.~Russakovsky, J.~Deng, H.~Su, J.~Krause, S.~Satheesh, S.~Ma, Z.~Huang,
  A.~Karpathy, A.~Khosla, M.~S. Bernstein, A.~C. Berg, and L.~Feifei,
\newblock ``{ImageNet} large scale visual recognition challenge,''
\newblock {\em International Journal of Computer Vision}, vol. 115, no. 3, pp.
  211--252, 2015.

\bibitem{selvaraju2017grad}
R.~R. Selvaraju, M.~Cogswell, A.~Das, R.~Vedantam, D.~Parikh, and D.~Batra,
\newblock ``{Grad-CAM}: {V}isual explanations from deep networks via
  gradient-based localization,''
\newblock in {\em International Conference on Computer Vision}, 2017, pp.
  618--626.

\bibitem{zhou2016learning}
B.~Zhou, A.~Khosla, A.~Lapedriza, A.~Oliva, and A.~Torralba,
\newblock ``Learning deep features for discriminative localization,''
\newblock in {\em Computer Vision and Pattern Recognition}, 2016, pp.
  2921--2929.

\bibitem{zhang2018adversarial}
X.~Zhang, Y.~Wei, J.~Feng, Y.~Yang, and T.~Huang,
\newblock ``Adversarial complementary learning for weakly supervised pbject
  localization,''
\newblock in {\em Computer Vision and Pattern Recognition}, 2018, pp.
  1325--1334.

\bibitem{singh2017hide}
K.~K. Singh and Y.~J. Lee,
\newblock ``Hide-and-seek: {F}orcing a network to be meticulous for
  weakly-supervised object and action localization,''
\newblock in {\em International Conference on Computer Vision}, 2017, pp.
  3544--3553.

\bibitem{zhang2018self}
X.~Zhang, Y.~Wei, G.~Kang, Y.~Yang, and T.~S. Huang,
\newblock ``Self-produced guidance for weakly-supervised object localization,''
\newblock in {\em European Conference on Computer Vision}, 2018, pp. 610--625.

\bibitem{xue2019danet}
H.~Xue, C.~Liu, F.~Wan, J.~Jiao, X.~Ji, and Q.~Ye,
\newblock ``{DANet}: {D}ivergent activation for weakly supervised object
  localization,''
\newblock in {\em International Conference on Computer Vision}, 2019, pp.
  6588--6597.

\bibitem{yin2020dual}
J.~Yin, S.~Zhang, D.~Chang, Z.~Ma, and J.~Guo,
\newblock ``Dual-attention guided dropblock module for weakly supervised object
  localization,''
\newblock in {\em International Conference on Pattern Recognition}, 2020.

\bibitem{wah2011cub}
C.~Wah, S.~Branson, P.~Welinder, P.~Perona, and S.~Belongie,
\newblock ``The caltech-ucsd birds-200-2011dataset,''
  \url{http://www.vision.caltech.edu/visipedia/cub-200-2011.html}, 2011.

\bibitem{howard2017mobilenets}
A.~G. Howard, M.~Zhu, B.~Chen, D.~Kalenichenko, W.~Wang, T.~Weyand,
  M.~Andreetto, and H.~Adam,
\newblock ``{MobileNets}: {E}fficient convolutional neural networks for mobile
  vision applications,''
\newblock {\em arXiv}, 2017.

\bibitem{voc2007}
``The {PASCAL} visual object classes challenge ({VOC}2007),''
  \url{http://www.pascal-network.org/challenges/VOC/voc2007/index.html}.

\bibitem{lin2014microsoft}
T.~Lin, M.~Maire, S.~J. Belongie, J.~Hays, P.~Perona, D.~Ramanan, P.~Dollar,
  and C.~L. Zitnick,
\newblock ``Microsoft {COCO}: {C}ommon objects in context,''
\newblock in {\em European Conference on Computer Vision}, 2014, pp. 740--755.

\bibitem{tian2019fcos}
Z.~Tian, C.~Shen, H.~Chen, and T.~He,
\newblock ``{FCOS}: {F}ully convolutional one-stage object detection,''
\newblock in {\em International Conference on Computer Vision}, 2019.

\bibitem{zhang2020bridging}
S.~Zhang, C.~Chi, Y.~Yao, Z.~Lei, and S.~Z. Li,
\newblock ``Bridging the gap between anchor-based and anchor-free detection via
  adaptive training sample selection,''
\newblock in {\em Computer Vision and Pattern Recognition}, June 2020.

\bibitem{lin2017feature}
T.~Lin, P.~Dollár, R.~Girshick, K.~He, B.~Hariharan, and S.~Belongie,
\newblock ``Feature pyramid networks for object detection,''
\newblock in {\em Computer Vision and Pattern Recognition}, 2017, pp. 936--944.

\bibitem{mmdetection}
K.~Chen, J.~Wang, J.~Pang, Y.~Cao, Y.~Xiong, X.~Li, S.~Sun, W.~Feng, Z.~Liu,
  J.~Xu, Z.~Zhang, D.~Cheng, C.~Zhu, T.~Cheng, Q.~Zhao, B.~Li, X.~Lu, R.~Zhu,
  Y.~Wu, J.~Dai, J.~Wang, J.~Shi, W.~Ouyang, C.-C. Loy, and D.~Lin,
\newblock ``{MMDetection}: Open mmlab detection toolbox and benchmark,''
\newblock {\em arXiv preprint arXiv:1906.07155}, 2019.

\bibitem{mmseg2020}
MMSegmentation Contributors,
\newblock ``{MMSegmentation}: Openmmlab semantic segmentation toolbox and
  benchmark,'' \url{https://github.com/open-mmlab/mmsegmentation}, 2020.

\bibitem{cordts2016cityscapes}
Ma. Cordts, M.~Omran, S.~Ramos, T.~Rehfeld, M.~Enzweiler, R.~Benenson,
  U.~Franke, S.~Roth, and B.~Schiele,
\newblock ``The cityscapes dataset for semantic urban scene understanding,''
\newblock in {\em Computer Vision and Pattern Recognition}, 2016.

\bibitem{sun2019deep}
K.~Sun, B.~Xiao, D.~Liu, and J.~Wang,
\newblock ``Deep high-resolution representation learning for human pose
  estimation,''
\newblock in {\em Computer Vision and Pattern Recognition}, 2019, pp.
  5693--5703.

\bibitem{maddox2019a}
W.~J. Maddox, P.~Izmailov, T.~Garipov, D.~Vetrov, and A.~G. Wilson,
\newblock ``A simple baseline for {B}ayesian uncertainty in deep learning,''
\newblock in {\em Neural Information Processing Systems}, 2019, pp.
  13153--13164.

\end{thebibliography}

\end{document}